\documentclass[twocolumn, switch]{article} 

\usepackage{preprint}

\usepackage{amsmath, amsthm, amssymb, amsfonts}
\usepackage{bm}
\usepackage{mathtools}

\usepackage[numbers,square]{natbib}
\usepackage[utf8]{inputenc}	
\usepackage[T1]{fontenc}	
\usepackage{xcolor}		
\usepackage[colorlinks = true,
            linkcolor = blue,
            urlcolor  = blue,
            citecolor = blue,
            anchorcolor = black]{hyperref}	
\usepackage{booktabs} 		
\usepackage{nicefrac}		
\usepackage{microtype}		
\usepackage{lineno}		
\usepackage{float}			

\usepackage{newfloat}
\DeclareFloatingEnvironment[name={Supplementary Figure}]{suppfigure}
\usepackage{sidecap}
\sidecaptionvpos{figure}{c}

\usepackage{titlesec}
\titlespacing\section{0pt}{12pt plus 3pt minus 3pt}{1pt plus 1pt minus 1pt}
\titlespacing\subsection{0pt}{10pt plus 3pt minus 3pt}{1pt plus 1pt minus 1pt}
\titlespacing\subsubsection{0pt}{8pt plus 3pt minus 3pt}{1pt plus 1pt minus 1pt}

\usepackage{graphics}
\usepackage{hyperref}
\usepackage{color}
\usepackage{multirow}
\usepackage{hhline}
\usepackage[flushleft]{threeparttable}
\usepackage{subfigure}

\newcommand{\etal}{\textit{et al}.}

\newcommand{\ie}{\textit{i}.\textit{e}.}
\newcommand{\eg}{\textit{e}.\textit{g}.}

\newcommand{\floor}[1]{\left\lfloor #1 \right\rfloor}

\title{Learned Image Downscaling for Upscaling using Content Adaptive Resampler}

\usepackage{authblk}

\author{Wanjie Sun}
\author{Zhenzhong Chen\thanks{\tt{zzchen@ieee.org}}}
\affil{School of Remote Sensing and Information Engineering, Wuhan University}

\begin{document}

\twocolumn[ 
  \begin{@twocolumnfalse} 
  
\maketitle

\begin{abstract}
Deep convolutional neural network based image super-resolution (SR) models have shown superior performance in recovering the underlying high resolution (HR) images from low resolution (LR) images obtained from the predefined downscaling methods. In this paper we propose a learned image downscaling method based on content adaptive resampler (CAR) with consideration on the upscaling process. The proposed resampler network generates content adaptive image resampling kernels that are applied to the original HR input to generate pixels on the downscaled image. Moreover, a differentiable upscaling (SR) module is employed to upscale the LR result into its underlying HR counterpart. By back-propagating the reconstruction error down to the original HR input across the entire framework to adjust model parameters, the proposed framework achieves a new state-of-the-art SR performance through upscaling guided image resamplers which adaptively preserve detailed information that is essential to the upscaling. Experimental results indicate that the quality of the generated LR image is comparable to that of the traditional interpolation based method and the significant SR performance gain is achieved by deep SR models trained jointly with the CAR model. The code is publicly available on: https://github.com/sunwj/CAR.
\end{abstract}
\vspace{0.35cm}

  \end{@twocolumnfalse} 
] 



\section{Introduction}
{\let\thefootnote\relax\footnote{{This work was supported in part by National Natural Science Foundation of China under contract No. 61771348. (Corresponding author: Zhenzhong Chen, E-mail:  \texttt{zzchen@ieee.org})}}}As the smartphone cameras are starting to rival or beat DSLR cameras, a large number of ultra high resolution images are produced everyday. However, it is always reduced from its original resolution to smaller sizes that are fit to the screen of different mobile devices and web applications. Thus, it is desirable to develop efficient image downscaling and upscaling method to make such application more practical and resources saving by only generating, storing and transmitting a single downscaled version for preview and upscaling it to high resolution when details are going to be viewed. Besides, the pre-downscaling and post-upscaling operation also helps to save storage and bandwidth for image or video compression and communication \cite{Bruckstein:2003,Lin:2006,zhang:2011,li:2019}.
	
Image downscaling is one of the most common image processing operations, aiming to reduce the resolution of the high-resolution (HR) image while keeping its visual appearance. According to the Nyquist-Shannon sampling theorem \cite{shannon:1998}, it is inevitable that high-frequency content will get lost during the sample-rate conversion. Contrary to the image downscaling task is the image upscaling, also known as resolution enhancement or super-resolution (SR), trying to recover the underlying HR image of the LR input. Image SR is essentially an ill-posed problem because an undersampled image could refer to numerous HR images. The quality of the SR result is very limited due to the ill-posed nature of the problem and the lost frequency components cannot be well-recovered \cite{yang:2010,yangjc:2017}. Previous work regards image downscaling and SR as independent tasks. Image downscaling techniques pay much attention to enhance details, such as edges, which helps to improve human visual perception \cite{zhang:2011}. On the other hand, recent state-of-the-art deep SR models have witnessed their capability to restore HR image from the LR version downscaled using traditional filtering-decimation based methods with great performance gain \cite{yang:2018,anwar:2019}. However, the predetermined downscaling operations may be sub-optimal to the SR task and state-of-the-art deep SR models still cannot well recover fine details from distorted textures caused by the fixed downscaling operations.

In this paper, we proposed a learned content adaptive image downscaling model in which a SR model tries to best recover HR images while adaptively adjusting the downscaling model to produce LR images with potential detailed information that are key to the optimal SR performance. The downscaling model is trained without any LR image supervision. To make sure that the LR image produced by our downscaling model is a valid image, we propose to employ the resampling method where content adaptive non-uniform resampling kernels predicted by a convolutional neural network (CNN) are applied to the original HR image to generate pixels of the LR output. Quantitative and qualitative experimental results illustrate that LR images produced by the proposed model can maintain comparable visual quality as the widely used bicubic interpolation based image downscaling method while advanced SR image quality is obtained using state-of-the-art deep SR models trained with LR images generated from the proposed CAR model.

Our contributions are concluded as follows:
\begin{itemize}
	\item A learned image downscaling model is proposed which is trained under the guidance of the SR model. The proposed image downscaling model produces images that can be well super-resolved while comparable visual quality can be maintained. Experimental results indicate a new state-of-the-art SR performance with the proposed end-to-end image downscaling and upscaling framework.
	\item The resampling method is employed to downscale image by applying content adaptive non-uniform resampling kernels on the original HR input, which can effectively maintain the structure of the HR input in an unsupervised manner. Because directly predicting the LR image by combining low and high-level abstract image features can not guarantee that the generated result is a genuine image without any LR image supervision.
	\item The learned content adaptive non-uniform resampling kernels perform non-uniform sampling and also make the size of resampling kernels to be more effective. The generated kernels produced by the proposed CAR model are composed of weights and sampling position offsets in both horizontal and vertical directions, making the learned resampling kernels adaptively change their weights and shape according to its corresponding resampling contents.
\end{itemize}
	
	The rest of the paper is organized as follows. Section \ref{sec:related_work} reviews task independent and task driven image downscaling algorithms. Section \ref{sec:model_arch} introduces the proposed SR guided content adaptive image downscaling framework, and computing process of each component in the framework is explained. Section \ref{sec:experiment} evaluates and analyzes experimental results for the SR images and downscaled images quantitatively and qualitatively. Finally, Section \ref{sec:conclusion} summarizes our work.
	
	\section{Related Work}\label{sec:related_work}
	This section presents a review about image downscaling techniques aiming to maintain the visual quality of the LR image. Image downscaling algorithms can be categorized into two groups as follows.
	
	\subsection{Task independent image downscaling} Earlier work of image downscaling primarily tends to prevent aliasing \cite{shannon:1998} which arises during sampling rate reduction. Those methods are based on linear filters \cite{wolberg:1990}, where the HR image is firstly convolved with a low-pass kernel to push frequency componenets of the image below the Nyquist frequency \cite{fang:2012}, then being sub-sampled into target size. Many frequency-based filters are developed, \eg, the box, bilinear and bicubic filter \cite{mitchell:1988}. However, the downscaled images tend to be blurred because the high-frequency details are suppressed. Filters that are designed to model the ideal \textit{sinc} filter, \eg, the Lanczos filter \cite{duchon:1979}, tend to produce ringing artifacts near strong image edges. All of these filters are predetermined with some of them having tuning parameters. The same filter is applied globally to the input HR image, ignoring characteristics of image content with varying details.
	
	Recently, many researchers begin to focus on the aspects of detail preserving and human perception when developing image downscaling algorithm. Kopf \etal  firstly proposed a novel content adaptive image downscaling method based on a joint bilateral filter \cite{kopf:2013}. The key idea is to optimize the shape and locations of the downsampling kernels to better align with local image features by considering both spatial and color variances of the local region. \"{O}ztireli and Gross \cite{oztireli:2015} proposed a method to downscale HR images without filtering. They consider image downscaling as an optimization problem and use the structural similarity index (SSIM) \cite{wang:2004} as objective to directly optimize the downscaled image against its original image. This approach helps to capture most of the perceptually important details. Weber \etal \cite{weber:2016} proposed an image downscaling algorithm aiming to preserve small details of the input image, which are often crucial for a faithful visual impression. The intuition is that small details transport more information than bigger areas with similar colors. To that end, an inverse bilateral filter is used to emphasize differences rather than punishing them. Gastal and Oliveira \cite{gastal:2017} introduced the spectral remapping algorithm to control aliasing during image downscaling. Instead of discarding high-frequency information, it remaps such information into the representable range of the downsampled spectrum. Recently, Liu \etal \cite{liu:2018} proposed a L0-regularized optimization framework for image downscaling, which is composed of a gradient-ratio prior and reconstruction prior. The downscaling problem is solved by iteratively optimize the two priors in an alternative way.
	
	\subsection{Task specific image downscaling} Most image downscaling algorithms only care about the visual quality of the downscaled image, so that the downscaled image may not be optimal to other computer vision tasks. To tackle this problem, task guided image downscaling has emerged. Zhang \etal \cite{zhang:2011} took the quality of the interpolated image from the downscaled counterpart into consideration. They proposed an interpolation-dependent image downscaling algorithm by modeling the downscaling operation as the inverse operator of upsampling. Benefiting from the well established deep learning frameworks, Hou \etal \cite{hou:2018} proposed a deep feature consistency network that is applicable to image mapping problems. One of the applications illustrated in the paper is image downscaling. The image downscaling network is trained by keeping the deep features of the input HR image and resulting LR image consistent through another pre-trained deep CNN. Kim \etal \cite{kim:2018} presented a task aware image downscaling model based on the auto-encoder and the bottleneck layer outputs the downscaled image. In their framework, the encoder acts as the image downscaling network and the decoder is the SR network. The task aware downscaled image is obtained by jointly training the encoder and decoder to maximize the SR performance. Similar to the framework presented by \cite{kim:2018}, Li \etal \cite{li:2019} proposed a convolutional neural network for image compact resolution named CNN-CR, which is composed of a CNN to estimate the LR image and a learned or specified upscaling model to reconstruct the HR image. The generative nature of the encoder like networks implicitly require additional information to constrain the output to be a valid image whose content resembles the HR image. In \cite{hou:2018}, in order to compute feature consistency loss against the HR image, they upsample the downscaled image back to the same size as the HR input using nearest neighbor interpolation. In \cite{kim:2018,li:2019}, guidance images, obtained using bicubic downsampling, are employed to constrain the output space of the LR image generating networks.
	
	\begin{figure*}[h]
		\centering
		\includegraphics[width=\textwidth]{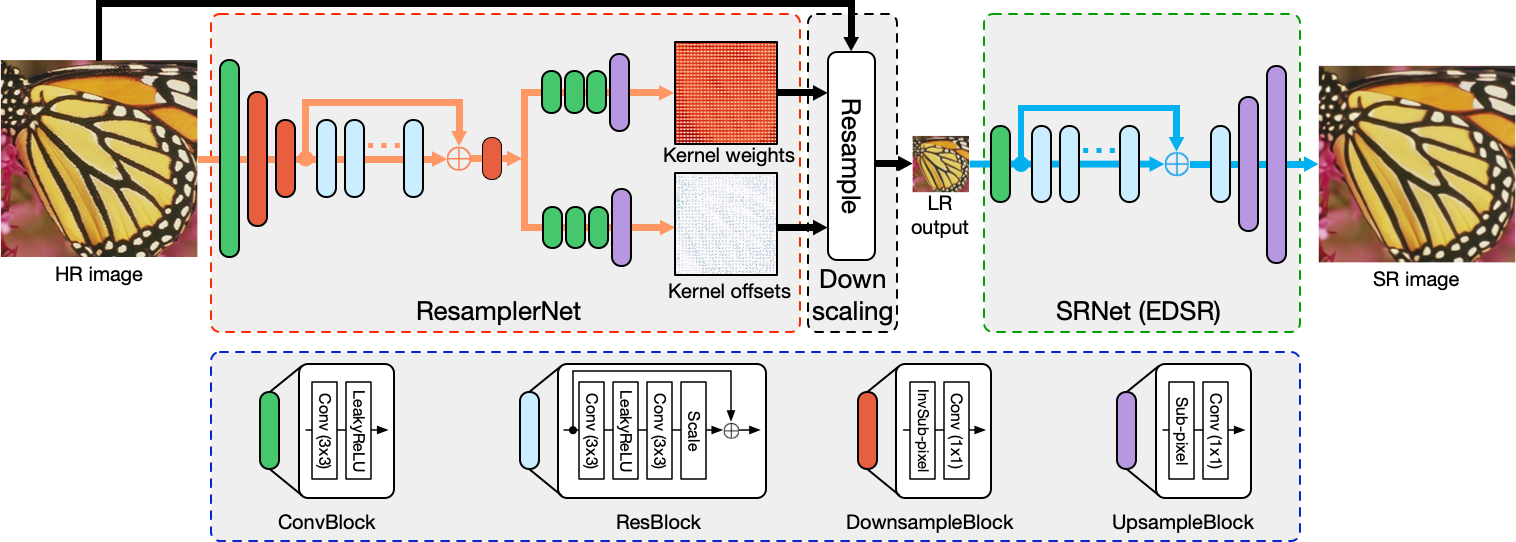}
		\caption{Network architecture. It consists of three parts, the ResamplerNet, the Downscaling module, and the SRNet. The ResamplerNet is designed to estimate content adaptive resampling kernels and its corresponding offsets for each pixel in the downscaled image. The SRNet, can be any form of differentiable upsampling operations, is employed to guide the training of the ResamplerNet by simply minimizing the SR error. The entire framework is trained end-to-end by back-propagating error signals through a the differentiable downscaling module. The composition of each building block is detailed on the blue dashed frame.}
		\label{fig:overview}
	\end{figure*}
	\section{Model Architecture}\label{sec:model_arch}
	This section introduces the architecture and formulation of the content adaptive resampler (CAR) model for image downscaling. As shown in Fig. \ref{fig:overview}, the framework is composed of two major components, \ie, the resampler generation network (ResamplerNet) and the SR network (SRNet). The ResamplerNet is responsible for estimating the content adaptive resampling kernels according to its input HR image, later the resampling kernels are applied to the input HR image to produce the downscaled image. The SRNet takes the resulting downscaled image as input and tries to restore the underlying HR image. The entire framework is trained end-to-end in an unsupervised manner where the primary objective we need to optimize is the SR reconstruction error with respect to the input HR image. By back-propagating the gradient of the reconstruction error through the SRNet and ResamplerNet, the gradient descent algorithm adjusts the parameters of the resampler generation network to produce better resampling kernels which make the downscaled image can be super-resolved more easily.
	
	\subsection{Content adaptive image downscaling resampler}
	We design the proposed content adaptive image downscaling model that is trained using the unsupervised strategy. Methods presented in \cite{hou:2018,kim:2018,li:2019} synthesize the downscaled image by combining latent representations of the HR image extracted by the CNN and proper constraints are required to make sure that the result is a meaningful image. In this paper, we propose to obtain the downscaled image using the idea of resampling the HR image, which effectively makes the downscaled result look like the original HR image without any constraints.
	
	The filters for traditional bilinear or bicubic downscaling are basically fixed, with the only variation being the shift of the kernel according to the location of the newly created pixel in the downscaled image. Contrary to this, we propose to use dynamic downscaling filters inspired by the dynamic filter networks \cite{DeBrabandere:2016}. The downscaling kernels are generated for each pixel in the downscaled image depending on the effective resampling region on the HR image. It can be considered as one type of meta-learning \cite{schmidhuber:1987} that learns how to resample. However, filter-based image resampling methods generally require a certain minimum kernel size to be effective \cite{liu:2018}. We alleviate this issue by taking the idea from the deformable convolutional networks \cite{dai:2017}. In addition to estimating the content adaptive resampling kernel weights, we also associate spatial offset with each element in the resampling kernel. The content adaptive resampling kernels with position offsets can be considered as learnable dilated (atrous) convolutions \cite{fisher} with the learned dilation rate. Besides, the offset for each kernel element can be different in both magnitude and direction, it can perform non-uniform sampling according to the content structure of the input HR image.
	
	We use a convolutional neural network with residual connections \cite{he:2016} to estimate the weights and offsets for each resampling kernel. The ResamplerNet consists of downscaling blocks, residual blocks and upscaling blocks. The downscaling and residual blocks are trained to model the context of the input HR image as a set of feature maps. Then, two upscaling blocks are used to learn the content adaptive resampling kernel weights $\mathbf{K} \in \mathbb{R}^{(h/s)\times (w/s)\times m\times n}$, offsets in the horizontal direction $\Delta\mathbf{Y} \in \mathbb{R}^{(h/s)\times (w/s)\times m\times n}$ and offsets in the vertical direction $\Delta\mathbf{X} \in \mathbb{R}^{(h/s)\times (w/s)\times m\times n}$, respectively. $h$ and $w$ are the height and width of the input HR image, $s$ is the downscaling factor, and $m$, $n$ represent the size of the content adaptive resampling kernel. Each kernel is normalized so that elements of a kernel are summed up to be $1$.
	
	\subsection{Image downscaling}
	\begin{figure}[h]
		\centering
		\includegraphics[width=0.6\columnwidth]{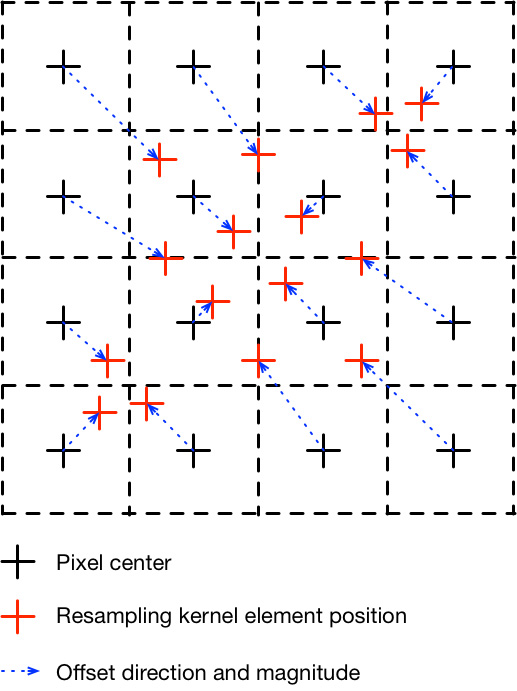}
		\caption{Non-uniform resampling. The black $+$ represents the center of a pixel in the HR image and the red $+$ indicate the position of the resampling kernel element. The blue dashed arrow shows the offset direction and magnitude of the resampling kernel element relative to its corresponding pixel center.}
		\label{fig:resampling}
	\end{figure}
	The estimated content adaptive resampling kernels are applied to the corresponding positions of the input HR image to construct the pixel in the downscaled image. For each output pixel, the same resampling kernel is simultaneously applied to three channels of the RGB image. As illustrated by Fig. \ref{fig:resampling}, pixels covered by the resampling kernel are weights summed to obtain the pixel value in the downscaled image.
	
	\textbf{Forward pass.} Firstly, we need to position each resampling kernel, associated with the pixel of the downscaled image, on the HR image. It can be achieved using the projection operation defined as:
	\begin{equation}\label{eqn:projection}
	(u, v) = (x + 0.5, y + 0.5) \times scale - 0.5
	\end{equation}
	where $(x, y)$ is the indices of a pixel at the $x$-th row and $y$-th column in the downscaled image, $scale$ represents the downscaling factor, and the resulting $(u, v)$ is the center of downscaled pixel $\bm{p}^\text{LR}_{x,y}$ projected on the input HR image. Equation \ref{eqn:projection} assumes that pixels have a nonzero size, and the distance between two neighboring samples is one pixel.
	
	Then, each pixel in the downscaled image is created by local filtering on pixels in the input HR image with the corresponding content adaptive resampling kernel as follows:
	\begin{equation}\label{equ:resampling}
	\begin{aligned}
	\bm{p}^\text{LR}_{x,y} &= \sum_{i=0}^{m-1}\sum_{j=0}^{n-1} \mathbf{K}_{x,y}(i, j) \cdot \bm{s}^\text{HR}\left(u + i - \frac{m}{2} + \Delta\mathbf{X}_{x,y}(i, j),\right.\\
	&\left.v + j - \frac{n}{2} + \Delta\mathbf{Y}_{x,y}(i, j)\right)
	\end{aligned}
	\end{equation}
	where $\mathbf{K}_{x,y} \in \mathbb{R}^{m\times n}$ is the resampling kernel associated with the downscaled pixel at location $(x, y)$, $\Delta\mathbf{X}_{x,y} \in \mathbb{R}^{m\times n}$ and $\Delta\mathbf{Y}_{x,y} \in \mathbb{R}^{m\times n}$ are the spatial offset for each element in the $\mathbf{K}_{x,y}$. The $\bm{s}^\text{HR}$ is the sample point value of the input HR image. Due to the location projection and fractional offsets, $\bm{s}^\text{HR}$ can refer to non-lattice point on the HR image, therefore, $\bm{s}^\text{HR}$ is computed by bilinear interpolating the nearby four pixels around the fractional position:
	\begin{equation}
	\begin{aligned}
	\bm{s}^\text{HR}_{u',v'} &= (1-\alpha)(1-\beta)\cdot\bm{p}^\text{HR}_{\floor{u'}, \floor{v'}} + \alpha(1 - \beta)\cdot\bm{p}^\text{HR}_{\floor{u'}, \floor{v'}+1}\\
	& + (1-\alpha)\beta\cdot\bm{p}^\text{HR}_{\floor{u'}+1, \floor{v'}} + \alpha\beta\cdot\bm{p}^\text{HR}_{\floor{u'}+1, \floor{v'}+1}
	\end{aligned}
	\end{equation}
	where $u'$ and $v'$ are fractional position on the HR image, $\alpha=u'-\floor{u'}$ and $\beta=v'-\floor{v'}$ are the bilinear interpolation weights.
	
	\textbf{Backward pass.} The ResamplerNet is trained using the gradient descent technique and we need to back-propagate gradients from the SRNet through the resampling operation. The partial derivative of the downscaled pixel with respect to the resampling kernel weight can be formulated as:
	\begin{equation}\label{eqn:kernel_gradient}
	\begin{aligned}
	\frac{\partial \bm{p}^\text{LR}_{x, y}}{\partial \mathbf{K}_{x,y}(i, j)} &= \bm{s}^\text{HR}\left(u + i - \frac{m}{2} + \Delta\mathbf{X}_{x,y}(i, j),\right.\\
	&\left.v + j - \frac{n}{2} + \Delta\mathbf{Y}_{x,y}(i, j)\right)
	\end{aligned}
	\end{equation}
	the partial derivative of downscaled pixel with respect to the element in the resampling kernel is simply the interpolated pixel value. Equation \ref{eqn:kernel_gradient} is derived with a single channel image and can be generalized to the color image by summing up values calculated separately on the R, G and B channels.
	
	The partial derivative of downscaled pixel with respect to the kernel element offset is computed as:
	\begin{equation}
	\begin{aligned}
	\frac{\partial \bm{p}^\text{LR}_{x, y}}{\partial \Delta\mathbf{X}_{x,y}(i,j)} &= \frac{\partial \bm{p}^\text{LR}_{x, y}}{\partial \bm{s}_{u',v'}^\text{HR}}\cdot\frac{\partial \bm{s}_{u',v'}^\text{HR}}{\partial \Delta\mathbf{X}_{x,y}(i,j)}\\
	\frac{\partial \bm{p}^\text{LR}_{x, y}}{\partial \Delta\mathbf{Y}_{x,y}(i,j)} &= \frac{\partial \bm{p}^\text{LR}_{x, y}}{\partial \bm{s}_{u',v'}^\text{HR}}\cdot\frac{\partial \bm{s}_{u',v'}^\text{HR}}{\partial \Delta\mathbf{Y}_{x,y}(i,j)}
	\end{aligned}
	\end{equation}
	because we employ bilinear interpolation to compute $\bm{s}_{u',v'}^\text{HR}$, therefore, the partial derivative of downscaled pixel with respect to the kernel offset is defined as:
	\begin{equation}\label{eqn:offset_gradient}
	\begin{aligned}
	&\frac{\partial \bm{p}^\text{LR}_{x, y}}{\partial \Delta\mathbf{X}_{x,y}(i,j)} =\\ &\mathbf{K}_{x,y}(i,j)\cdot(1-\beta)\cdot\left(\bm{p}^\text{HR}_{\floor{u'}, \floor{v'}+1} - \bm{p}^\text{HR}_{\floor{u'}, \floor{v'}}\right) +\\
	&\beta\cdot\left(\bm{p}^\text{HR}_{\floor{u'}+1, \floor{v'}+1} - \bm{p}^\text{HR}_{\floor{u'}+1, \floor{v'}}\right)\\
	&\frac{\partial \bm{p}^\text{LR}_{x, y}}{\partial \Delta\mathbf{Y}_{x,y}(i,j)} =\\ &\mathbf{K}_{x,y}(i,j)\cdot(1-\alpha)\cdot\left(\bm{p}^\text{HR}_{\floor{u'}+1, \floor{v'}} - \bm{p}^\text{HR}_{\floor{u'}, \floor{v'}}\right) +\\
	&\alpha\cdot\left(\bm{p}^\text{HR}_{\floor{u'}+1, \floor{v'}+1} - \bm{p}^\text{HR}_{\floor{u'}, \floor{v'}+1}\right)
	\end{aligned}
	\end{equation}
	also because Equation \ref{eqn:loss_offset} is defined with a single color image, we need to sum up partial derivative of each color component with respect to the offset to obtain the final partial derivative of downscaled pixel with respect to the kernel element offset.
	
	The pixel value of the downscaled image created using Equation \ref{equ:resampling} is inherently continuous floating point number. However, common image representation describes the color using integer number ranging from 0 to 255, thus, a quantization step is required. Since simply rounding the floating point number to its nearest integer number is not differentiable, we utilize the soft round method proposed by Nakanishi \etal\cite{nakanishi:2018} to derive the gradient in the back-propagation during the training phase. The soft round function is defined as:
	\begin{equation}
	\text{round}_\text{soft}(x) = x - \alpha\frac{\sin{2\pi x}}{2\pi}
	\end{equation}
	where $\alpha$ is a tuning parameter used to adjust the gradient around the integer position. Note that in the forward propagation, the non-differentiable round function is used to get the nearest integer value.
	
	\subsection{Image upscaling}
	The proposed CAR model is trained using back-propagation to maximize the SR performance. The image upscaling module can be of any form of SR networks, even the differentiable bilinear or bicubic upscaling operations. After the seminal super-resolution model using deep learning proposed by Dong \etal \cite{dong:2016}, \ie, the SRCNN, many state-of-the-art neural SR models have been proposed \cite{kim:2016,kim:2016-2,shi:2016,ledig:2017,lim:2017,tong:2017,zhang:2018,zhanglei:2018,zhang:2018eccv,lai:2018,zkzl:2019,dai:2019}. More reviews and discussion about single image SR using deep learning can be referred to \cite{yang:2018}. This paper employs state-of-the-art SR model EDSR \cite{lim:2017} as the image upscaling module to guide the training of the proposed CAR model.
	
	The EDSR is one of the state-of-the-art deep SR networks and its superior performance is benefited from the powerful residual learning techniques \cite{he:2016}. The EDSR is built on the success of the SRResNet \cite{ledig:2017} which achieved good performance by simply employing the ResNet \cite{he:2016} architecture on the SR task. The EDSR enhanced SR performance by removing unnecessary parts (Batch Normalization layer \cite{Ioffe:2015}) from the SRResNet and also applying a number of tweaks, such as adding residual scaling operation to stable the training process \cite{lim:2017}. The SRNet part of Fig. \ref{fig:overview} shows the architecture of the EDSR. It is composed of convolution layers converting the RGB image into feature spaces and a group of residual blocks refining feature maps. A global residual connection is employed to improve the efficiency of the gradient back-propagation. Finally, sub-pixel layers are utilized to upsample and transform features into the target SR image.
	
	\subsection{Training objectives}\label{sec:train_objective}
	One of the main contributions of our work is that we propose a model to learn image downscaling without any supervision signifing that no constraint is applied to the downscaled image. The only objective guiding the generation of the downscaled image is the SR restoration error. The most common loss function generally defaults to the SR network training is the mean squared error (MSE) or the L2 norm loss, but it tends to lead to poor image quality as perceived by human observers \cite{zhao:2017}. Lim \etal \cite{lim:2017} found that using another local metric, \eg, L1 norm, can speed up the training process and produce better visual results. In order to improve the visual fidelity of the super-resolved image, perceptually-motivated metrics, such as SSIM \cite{wang:2004}, MS-SSIM \cite{wang:2003} and perceptual loss \cite{justin:2016} are usually incorporated in the SR network training. To do fair comparisons with the EDSR, we only adopt the L1 norm loss as the restoration metric as suggested by \cite{lim:2017}. The L1 norm loss defined for SR is:
	\begin{equation}
	\mathcal{L}^{\text{L}_1}(\hat{\mathbf{I}}) = \frac{1}{N}\sum_{\bm{p}\in\mathbf{I}}|\bm{p} - \hat{\bm{p}}|
	\end{equation}
	where $\hat{\mathbf{I}}$ is the SR result, $\bm{p}$ and $\hat{\bm{p}}$ represent the ground-truth and reconstructed pixel value, $N$ indicates the number of pixels times the number of color channels.
	
	We associate spatial offset for each element in the resampling kernel, and the offset is estimated without taking the neighborhoods of the kernel element into account. Independent kernel element offset may break the topology of the resampling kernel. To alleviate this problem, we suggest using the total offset distance of all kernel elements as a regularization which encourages kernel elements to stay in their rest position (avoid unnecessary movements). Additionally, since the pixels indexed by the kernel elements that are far from the central position may have less correlation to the resampling result, we assign different weights to their corresponding offset distance in terms of their position relative to the central position. The offset distance regularization term for a single resampling kernel is thus formulated as:
	\begin{equation}\label{eqn:loss_offset}
	\mathcal{L}^\text{offset}_{x,y} = \sum_{i=0}^{m-1}\sum_{j=0}^{n-1}\eta + \sqrt{\Delta\mathbf{X}_{x,y}(i,j)^2 + \Delta\mathbf{Y}_{x,y}(i,j)^2}\cdot w(i, j)
	\end{equation}
	where $w(i, j) = \sqrt{(i-\frac{m}{2})^2 + (j-\frac{n}{2})^2} \mathbin{/} \sqrt{\frac{m}{2} ^ 2 + \frac{n}{2}^2}$ is the normalized distance weight, the $\eta$ is introduced to act as the offset distance weight regulator.
	
	The inconsistent resampling kernel offset of spatially neighboring resampling kernels may cause pixel phase shift on the resulting LR images, which is manifested as jaggies, especially on the vertical and horizontal sharp edges (\eg, the LR image in Fig. \ref{fig:aliasing_optimized} (b)). To alleviate this phenomenon, we introduce the total variation (TV) loss \cite{RUDIN:1992} to constrain the movement of spatially neighboring resampling kernels. Instead of constraining the offsets on both vertical and horizontal directions, we only regularize vertical offsets on the horizontal direction and horizontal offsets on the vertical direction, which we name it the partial TV loss. Besides, variations of each resampling kernel offsets are weighted by its corresponding resampling kernel weights, leading to the following formula:
	\begin{equation}
	\begin{aligned}
	\mathcal{L}^{\text{TV}} &= \sum_{x,y}\left(\sum_{i,j}|\Delta\mathbf{X}_{\cdot, y+1}(i,j) - \Delta\mathbf{X}_{\cdot, y}(i,j)| \cdot \mathbf{K}(i, j)\right.\\
	&\left.+ \sum_{i,j}|\Delta\mathbf{Y}_{x+1, \cdot}(i,j) - \Delta\mathbf{Y}_{x, \cdot}(i,j)| \cdot \mathbf{K}(i, j)\right)
	\end{aligned}
	\end{equation}
	
	Finally, the optimization objective of the entire framework is defined as:
	\begin{equation}
	\mathcal{L} = \mathcal{L}^{\text{L}_1} + \lambda\overline{\mathcal{L}^\text{offset}} + \gamma\mathcal{L}^{\text{TV}}
	\end{equation}
	where the $\overline{\mathcal{L}^\text{offset}}$ is the mean offset distance regularization term of all the resampling kernels, and $\lambda$ is a scalar introduced to control the strength of offset distance regularization. $\gamma$ is also a scalar used to tune the contribution of the partial TV loss to the final optimization objective.
	
	\section{Experiments}\label{sec:experiment}
	\subsection{Experimental setup}
	\subsubsection{Datasets and metrics}
	For training the proposed content adaptive image downscaling resampler generation network under the guidance of EDSR, we employed the widely used DIV2K \cite{agustsson:2017} image dataset. There are 1000 high-quality images in the DIV2K dataset, where 800 images for training, 100 images for validation and the other 100 images for testing. In the testing, four standard datasets, \ie, the Set5 \cite{bevilacqua:2012}, Set14 \cite{roman:2012}, BSD100 \cite{martin:2001} and Urban100 \cite{huang:2015} were used as suggested by the EDSR paper \cite{lim:2017}. Since we focus on how to downscale images without any supervision, only HR images of the mentioned datasets were utilized. Following the setting in \cite{lim:2017}, we evaluated the peak noise-signal ratio (PSNR) and SSIM \cite{wang:2004} on the Y channel of images represented in the YCbCr (Y, Cb, Cr) color space.
	
	\subsubsection{Implementation details}\label{sec:implementation_details}
	Regarding the implementation of the ResamplerNet, we first subtracted the mean RGB value of the DIV2K training set. During the downsampling process, we gradually increased the channels of the output feature map from 3 to 128 using $3\times 3$ convolution operation followed by the \texttt{LeakyReLU} activation. 5 residual blocks with each having features of 128 channels are used to model the context. For the two branches estimating the resampling kernels and offsets, we used the same architecture which is composed of `\texttt{Conv}-\texttt{LeakyReLU}' pairs with 256 feature channels. A sub-pixel convolution was applied to upscale and transform the input feature maps into resampling kernels and offsets. For the configuration of the EDSR, we adopted the one with 32 residual blocks and 256 features for each convolution in the residual block. 
	
	One of the important hyper-parameters must be determined is the resampling kernel size and the unit offset length. We defined a $3\times 3$ kernel size on the downscaled image space. Its actual size on the HR image space is $(3\times s) \times (3\times s)$, where $s$ is the downscaling factor. The unit offset length was defined as one pixel on the downscaled image space whose corresponding unit length on the HR image space is $s$. For the offset distance weight regulator in Equation \ref{eqn:loss_offset}, we empirically set it to be $1$.
	
	The entire framework was trained on the DIV2K training set using the Adam optimizer \cite{kingma:2015} with $\beta_1=0.9$, $\beta_2=0.999$ and $\epsilon=10^{-6}$. We set the mini-batch size as 16, and randomly crop the input HR image into $192\times 192$ (for $4\times$ downscale and SR) and $96\times 96$ (for $2\times$ downscale and SR) patches. Training samples were augmented by applying random horizontal and vertical flip. During training, we conducted validation using 10 images from the DIV2K validation set to select the trained model parameters, and the PSNR on validation was performed on full RGB channels \cite{lim:2017}. The initial learning rate was $10^{-4}$ and decreased when the validation performance does not increase within 100 epoch.
	
	\subsection{Evaluation of downscaling methods for SR}
	This section reports the quantitative and qualitative performance of different image downscaling methods for SR. Then results of ablation studies of the proposed CAR model is presented. We compared the CAR model with four image downscaling methods, \ie, the bicubic downscaling (Bicubic), and other three state-of-the-art image downscaling methods: perceptually optimized image downscaling (Perceptually) \cite{oztireli:2015}, detail-preserving image downscaling (DPID) \cite{weber:2016}, and L0-regularized image downscaling (L0-regularized) \cite{liu:2018}. We trained SR models using LR images downscaled by those four downscaling algorithms and LR images downscaled by the proposed CAR model. The DPID requires to manually tune a hyper-parameter, which is content variant, to produce better perceptually favorable results. However, it is unpractical for us to generate large amount LR images by manually tuning, also different people may have different perceptual preference. Thus, default value provided by the source code was adopted.
	
	\begin{table*}[h]
		\caption{Quantitative evaluation results (PSNR / SSIM) of different image downscaling methods for SR on benchmark datasets: Set5, Set14, BSD100, Urban100 and DIV2K (validation set).}\label{tab:comparision}
		\centering
		\resizebox{\textwidth}{!}{\begin{threeparttable}
			\begin{tabular}{c|c|c|c|c|c|c|c|c} 
				\hline
				\multicolumn{2}{c|}{Upscaling}                                                     & \multicolumn{2}{c|}{Bicubic}                                                                                                                             & \multicolumn{5}{c}{EDSR}                                                                                                                                                                                  \\ 
				\hline
				\multicolumn{2}{c|}{Downscaling}                                                   & Bicubic                                                                      & CAR                                                                       & Bicubic                                                                     & Perceptually   & DPID           & L0-regularized & \multicolumn{1}{c}{CAR}                                                 \\ 
				\hline\hline
				\multirow{2}{*}{Set5}                                                         & 2x & \textcolor[rgb]{0.016,0.2,1}{33.66} / \textcolor[rgb]{0.016,0.2,1}{0.9299}   & \textcolor[rgb]{1,0.122,0}{34.38~}/ \textcolor[rgb]{1,0.122,0}{0.9426}    & \textcolor[rgb]{0.016,0.2,1}{38.06} / \textcolor[rgb]{0.016,0.2,1}{0.9615}  & 31.45 / 0.9212 & 37.02 / 0.9573 & 35.40 / 0.9514 & \textcolor[rgb]{1,0.141,0}{38.94~}/ \textcolor[rgb]{1,0.141,0}{0.9658}   \\ 
				\cline{2-2}
				& 4x & \textcolor[rgb]{0.016,0.2,1}{28.42} / \textcolor[rgb]{0.016,0.2,1}{0.8104}   & \textcolor[rgb]{1,0.122,0}{28.93~}/ \textcolor[rgb]{1,0.122,0}{0.8335}    & \textcolor[rgb]{0.016,0.2,1}{32.35~}/ \textcolor[rgb]{0.016,0.2,1}{0.8981}  & 25.65 / 0.7805 & 31.75 / 0.8913 & 31.05 / 0.8847 & \textcolor[rgb]{1,0.141,0}{33.88~}/ \textcolor[rgb]{1,0.141,0}{0.9174}   \\ 
				\hline\hline
				\multirow{2}{*}{Set14}                                                        & 2x & \textcolor[rgb]{0.016,0.2,1}{30.24} / \textcolor[rgb]{0.016,0.2,1}{0.8688}   & \textcolor[rgb]{1,0.122,0}{31.01} / \textcolor[rgb]{1,0.122,0}{0.8908}    & \textcolor[rgb]{0.016,0.2,1}{33.88~}/ \textcolor[rgb]{0.016,0.2,1}{0.9202}  & 29.26 / 0.8632 & 32.82 / 0.9119 & 31.56 / 0.9008 & \textcolor[rgb]{1,0.141,0}{35.61~}/ \textcolor[rgb]{1,0.141,0}{0.9404}   \\ 
				\cline{2-2}
				& 4x & \textcolor[rgb]{0.016,0.2,1}{26.00} / \textcolor[rgb]{0.016,0.2,1}{0.7027}   & \textcolor[rgb]{1,0.122,0}{26.39~}~\textcolor[rgb]{1,0.122,0}{0.7326}     & \textcolor[rgb]{0.016,0.2,1}{28.64~}/ \textcolor[rgb]{0.016,0.2,1}{0.7885}  & 24.21 / 0.6684 & 28.27 / 0.7784 & 27.67 / 0.7702 & \textcolor[rgb]{1,0.141,0}{30.31} / \textcolor[rgb]{1,0.141,0}{0.8382}   \\ 
				\hline\hline
				\multirow{2}{*}{B100}                                                         & 2x & \textcolor[rgb]{0.016,0.2,1}{29.56} / \textcolor[rgb]{0.016,0.2,1}{0.8431}   & \textcolor[rgb]{1,0.122,0}{30.18~}/ \textcolor[rgb]{1,0.122,0}{0.8714}    & \textcolor[rgb]{0.016,0.2,1}{32.31~}/ \textcolor[rgb]{0.016,0.2,1}{0.9021}  & 28.62 / 0.8383 & 31.47 / 0.8922 & 30.75 / 0.8816 & \textcolor[rgb]{1,0.141,0}{33.83~}/ \textcolor[rgb]{1,0.141,0}{0.9262}   \\ 
				\cline{2-2}
				& 4x & \textcolor[rgb]{0.016,0.2,1}{25.96} / \textcolor[rgb]{0.016,0.2,1}{0.6675}   & \textcolor[rgb]{1,0.122,0}{26.17} / \textcolor[rgb]{1,0.122,0}{0.6963}    & \textcolor[rgb]{0.016,0.2,1}{27.71} / \textcolor[rgb]{0.016,0.2,1}{0.7432}  & 24.61 / 0.6391 & 27.27 / 0.7341 & 27.00 / 0.7293 & \textcolor[rgb]{1,0.141,0}{29.15~}/ \textcolor[rgb]{1,0.141,0}{0.8001}   \\ 
				\hline\hline
				\multirow{2}{*}{Urban100}                                                     & 2x & \textcolor[rgb]{0.016,0.2,1}{26.88 }/ \textcolor[rgb]{0.016,0.2,1}{0.8403}   & \textcolor[rgb]{1,0.122,0}{27.38~}/ \textcolor[rgb]{1,0.122,0}{0.8620}    & \textcolor[rgb]{0.016,0.2,1}{32.92~}/ \textcolor[rgb]{0.016,0.2,1}{0.9359}  & 26.39 / 0.8483 & 31.64 / 0.9271 & 30.23 / 0.9172 & \textcolor[rgb]{1,0.141,0}{35.24}~/ \textcolor[rgb]{1,0.141,0}{0.9572}   \\ 
				\cline{2-2}
				& 4x & \textcolor[rgb]{0.016,0.2,1}{23.14} / \textcolor[rgb]{0.016,0.2,1}{0.6577 }  & \textcolor[rgb]{1,0.122,0}{23.35~}/ \textcolor[rgb]{1,0.122,0}{0.6844}    & \textcolor[rgb]{0.016,0.2,1}{26.62~}/ \textcolor[rgb]{0.016,0.2,1}{0.8041}  & 21.58 / 0.6295 & 26.07 / 0.7967 & 25.83 / 0.7957 & \textcolor[rgb]{1,0.141,0}{29.28~}/ \textcolor[rgb]{1,0.141,0}{0.8711}   \\ 
				\hline\hline
				\multirow{2}{*}{\begin{tabular}[c]{@{}c@{}}DIV2K\\(validation) \end{tabular}} & 2x & \textcolor[rgb]{0.016,0.2,1}{31.01} / \textcolor[rgb]{1,0.137,0}{0.9393 }    & \textcolor[rgb]{1,0.122,0}{33.18~}/ \textcolor[rgb]{0.016,0.2,1}{0.9317}  & \textcolor[rgb]{0.016,0.2,1}{36.76} / \textcolor[rgb]{0.016,0.2,1}{0.9482}  & 31.23 / 0.8984 & 35.75 / 0.9419 & 34.69 / 0.9354 & \textcolor[rgb]{1,0.141,0}{38.26~}/~\textcolor[rgb]{1,0.145,0}{0.9599}   \\ 
				\cline{2-2}
				& 4x & \textcolor[rgb]{0.016,0.2,1}{26.66} / \textcolor[rgb]{0.016,0.2,1}{0.8521}   & \textcolor[rgb]{1,0.122,0}{28.50~}/ \textcolor[rgb]{1,0.122,0}{0.8557}    & \textcolor[rgb]{0.016,0.2,1}{31.04} / \textcolor[rgb]{0.016,0.2,1}{0.8452}  & 26.28 / 0.7381 & 30.53 / 0.8373 & 30.18 / 0.8340 & \textcolor[rgb]{1,0.141,0}{32.82~}/~\textcolor[rgb]{1,0.145,0}{0.8837}   \\
				\hline
			\end{tabular}
			\begin{tablenotes}
				\small
				\item Note: \textcolor{red}{Red} color indicates the best performance and \textcolor{blue}{Blue} color represents the second.
			\end{tablenotes}
		\end{threeparttable}}
	\end{table*}
	\subsubsection{Quantitative and qualitative analysis}
	Table \ref{tab:comparision} summarizes the quantitative comparison results of different image downscaling methods for SR. It consists of two parts, one for bicubic upscaling and one for upscaling using the EDSR. We first analyze the SR performance using the EDSR. As shown in Table \ref{tab:comparision} (Upscaling$\rightarrow$EDSR), the proposed CAR model trained under the guidance of EDSR considerably boosts the PSNR metric over all the testing cases, and a noticeable gain on the SSIM metric is also obtained. The significant performance gain is benefited from the joint training of the CAR image downscaling model and the EDSR in the end-to-end manner, where the goal of maximizing SR performance encourages the CAR to estimate better resamplers that produce the most suitable downscaled image for SR.
	
	When compared to the SR performance of the LR images downscaled by the bicubic interpolation, the three state-of-the-art image downscaling algorithms can hardly achieve satisfying results, although the visual quality superiority of the downscaled images is reported by those original work. This is because those image downscaling methods are designed for better human perception thus the original information is changed considerably, which makes the downscaled image not well adapted to the SR defined by distortion metrics. Additionally, compared to the SR baseline of the bicubic image downscaling, we note a significant performance degradation on the perceptually based image downscaling method. This indicates that downscaled image produced by the SSIM optimization cannot be well super-resolved by the state-of-the-art EDSR. The key reason can be illustrated as the SSIM optimization depends on patch selection which may lead to sub-pixel offset in the downscaled image. Other artifacts may underperform SR includes color splitting and noise exaggeration incurred during SSIM optimization \cite{liu:2018}.
	
	In addition, we also evaluated SR performance of the CAR model trained under the guidance of the bicubic interpolation based upscaling where the bicubic downscaling was used as the baseline. As reported in Table \ref{tab:comparision} (Upscaling$\rightarrow$Bicubic), the CAR model outperforms the fixed bicubic downscaling methods in terms of upscaling using the fixed bicubic interpolation. The comparison results demonstrate the effectiveness of the proposed CAR model that it is flexible and can be trained under the guidance of differentiable upscaling operations, even if the upscaling operator is not learnable. With this discovery, the proposed CAR model can potentially replace the traditional and commonly used bicubic image downscaling operation under the hood, and end users can obtain extra image zoom in quality gain freely when using the bicubic interpolation for upscaling.
	
	\begin{table*}[h]
		\caption{Evaluation results (PSNR / SSIM) of 4$\times$ upscaling using different SR networks on benchmark images downscaled by the CAR model.}\label{tab:more_comparision}
		\centering
		\resizebox{\textwidth}{!}{\begin{threeparttable}
			\begin{tabular}{c|c|c|c|c|c|c} 
				\hline
				Upscaling  & Downscaling & Set5 & Set14 & B100 & Urban100 & DIV2K                                                                        \\ 
				\hline\hline
				\multirow{3}{*}{SRDenseNet} & Bicubic & 32.02 / 0.8934 & 28.50 / 0.7782 & 27.53 / 0.7337 & 26.05 / 0.7819 & - / - \\ 
				\cline{2-2}
				& CAR\textdagger & \textcolor[rgb]{1,0.133,0}{33.16}/ \textcolor[rgb]{1,0.133,0}{0.9067} & \textcolor[rgb]{1,0.133,0}{29.85} / \textcolor[rgb]{1,0.133,0}{0.8201} & \textcolor[rgb]{1,0.133,0}{28.73} / \textcolor[rgb]{1,0.133,0}{0.7794} & \textcolor[rgb]{1,0.133,0}{27.97} / \textcolor[rgb]{1,0.133,0}{0.8403} & \textcolor[rgb]{1,0.133,0}{32.24} / \textcolor[rgb]{1,0.133,0}{0.8674} \\ 
				\cline{2-2}
				& CAR\textdaggerdbl & \textcolor[rgb]{0.016,0.2,1}{32.63} / \textcolor[rgb]{0.016,0.2,1}{0.9047}  & \textcolor[rgb]{0.016,0.2,1}{29.24} / \textcolor[rgb]{0.016,0.2,1}{0.8122}  & \textcolor[rgb]{0.016,0.2,1}{28.44} / \textcolor[rgb]{0.016,0.2,1}{0.7781}  & \textcolor[rgb]{0.016,0.2,1}{27.12} / \textcolor[rgb]{0.016,0.2,1}{0.8248}  & \textcolor[rgb]{0.016,0.2,1}{31.77} / \textcolor[rgb]{0.016,0.2,1}{0.8654}   \\ 
				\hline\hline
				\multirow{3}{*}{D-DBPN} & Bicubic & 32.47 / 0.8980 & 28.82 / 0.7860 & 27.72 / 0.7400 & 26.38 / 0.7946 & - / - \\ 
				\cline{2-2}
				& CAR\textdagger & \textcolor[rgb]{1,0.133,0}{33.07} / \textcolor[rgb]{1,0.133,0}{0.9061} & \textcolor[rgb]{1,0.133,0}{29.75} / \textcolor[rgb]{1,0.133,0}{0.8189} & \textcolor[rgb]{1,0.133,0}{28.70} / \textcolor[rgb]{1,0.133,0}{0.7789} & \textcolor[rgb]{1,0.133,0}{27.98} / \textcolor[rgb]{1,0.133,0}{0.8376} & \textcolor[rgb]{1,0.133,0}{32.13} / \textcolor[rgb]{1,0.133,0}{0.8664} \\ 
				\cline{2-2}
				& CAR\textdaggerdbl & \textcolor[rgb]{0.016,0.2,1}{32.71} / \textcolor[rgb]{0.016,0.2,1}{0.9055}  & \textcolor[rgb]{0.016,0.2,1}{29.17} / \textcolor[rgb]{0.016,0.2,1}{0.8076}  & \textcolor[rgb]{0.016,0.2,1}{28.45} / \textcolor[rgb]{0.016,0.2,1}{0.7784}  & \textcolor[rgb]{0.016,0.2,1}{27.00} / \textcolor[rgb]{0.016,0.2,1}{0.8222}  & \textcolor[rgb]{0.016,0.2,1}{31.76} / \textcolor[rgb]{0.016,0.2,1}{0.8650}   \\ 
				\hline\hline
				\multirow{3}{*}{RDN} & Bicubic & 32.47 / 0.8990 & 28.81 / 0.7871 & 27.72 / 0.7419 & 26.61 / 0.8028 & - / - \\ 
				\cline{2-2}
				& CAR\textdagger & \textcolor[rgb]{1,0.133,0}{33.34} / \textcolor[rgb]{1,0.133,0}{0.9132} & \textcolor[rgb]{1,0.133,0}{29.93} / \textcolor[rgb]{1,0.133,0}{0.8308} & \textcolor[rgb]{1,0.133,0}{28.89} / \textcolor[rgb]{1,0.133,0}{0.7961} & \textcolor[rgb]{1,0.133,0}{28.53} / \textcolor[rgb]{1,0.133,0}{0.8582} & \textcolor[rgb]{1,0.133,0}{32.32} / \textcolor[rgb]{1,0.133,0}{0.8756} \\ 
				\cline{2-2}
				& CAR\textdaggerdbl & \textcolor[rgb]{0.016,0.2,1}{33.15} / \textcolor[rgb]{0.016,0.2,1}{0.9112}  & \textcolor[rgb]{0.016,0.2,1}{29.59} / \textcolor[rgb]{0.016,0.2,1}{0.8227}  & \textcolor[rgb]{0.016,0.2,1}{28.79} / \textcolor[rgb]{0.016,0.2,1}{0.7913}  & \textcolor[rgb]{0.016,0.2,1}{27.69} / \textcolor[rgb]{0.016,0.2,1}{0.8412}  & \textcolor[rgb]{0.016,0.2,1}{32.20} / \textcolor[rgb]{0.016,0.2,1}{0.8747}   \\ 
				\hline\hline
				\multirow{3}{*}{RCAN} & Bicubic & 32.63 / 0.9002 & 28.87 / 0.7889 & 27.77 / 0.7436 & 26.82 / 0.8087 & 30.77 / 0.8459 \\ 
				\cline{2-2}
				& CAR\textdagger & \textcolor[rgb]{1,0.133,0}{33.84} / \textcolor[rgb]{1,0.133,0}{0.9187} & \textcolor[rgb]{1,0.133,0}{30.27} / \textcolor[rgb]{1,0.133,0}{0.8383} & \textcolor[rgb]{1,0.133,0}{29.16} / \textcolor[rgb]{1,0.133,0}{0.8021} & \textcolor[rgb]{1,0.133,0}{29.23} / \textcolor[rgb]{1,0.133,0}{0.8719} & \textcolor[rgb]{1,0.133,0}{32.81} / \textcolor[rgb]{1,0.133,0}{0.8842} \\ 
				\cline{2-2}
				& CAR\textdaggerdbl & \textcolor[rgb]{0.016,0.2,1}{33.37} / \textcolor[rgb]{0.016,0.2,1}{0.9138}  & \textcolor[rgb]{0.016,0.2,1}{29.87} / \textcolor[rgb]{0.016,0.2,1}{0.8294}  & \textcolor[rgb]{0.016,0.2,1}{28.95} / \textcolor[rgb]{0.016,0.2,1}{0.7953}  & \textcolor[rgb]{0.016,0.2,1}{28.28} / \textcolor[rgb]{0.016,0.2,1}{0.8541}  & \textcolor[rgb]{0.016,0.2,1}{32.46} / \textcolor[rgb]{0.016,0.2,1}{0.8786}   \\
				\hline
			\end{tabular}
			\begin{tablenotes}
				\small
				\item CAR\textdagger: the CAR model is trained jointly with its corresponding SR model.
				\item CAR\textdaggerdbl: the SR model is trained using the downscaled images generated by the CAR model that is jointly with the EDSR.
				\item Note: \textcolor{red}{Red} color indicates the best performance and \textcolor{blue}{Blue} color represents the second. The `-' indicates that results are not provided by the corresponding original publication.
			\end{tablenotes}
		\end{threeparttable}}
	\end{table*}
	To further validate the effectiveness of the CAR image downscaling model, we evaluated the CAR model trained with another four state-of-the-art deep SR models, \ie, the SRDenseNet \cite{tong:2017}, D-DBPN \cite{haris:2018}, RDN \cite{zhang:2018} and RCAN \cite{zhang:2018eccv}, using 4$\times$ downscaling and upscaling factor on five testing datasets. We trained all models using the DIV2K training dataset and all other training setup is set to be the same as described in Section \ref{sec:implementation_details}. Table \ref{tab:more_comparision} presents the PSNR and SSIM of 4$\times$ upscaled images corresponding to LR images generated using the bicubic interpolation (MATLAB's \texttt{imresize} function with default settings) and the CAR model. Experimental results (Bicubic and CAR\textdagger) demonstrate a consistent performance gain of the SR task on images downscaled using the CAR model against that of the bicubic interpolation downscaling. When considering the SR performance of the CAR model trained under the guidance of the bicubic interpolation (Table \ref{tab:comparision}) we can reasonably arrive at the conclusion that the CAR image downscaling model can be learned to adapt to SR models as long as the SR operation is differentiable.
	
	In order to illustrate that the CAR image downscaling model can effectively preserve essential information which can help SR models learn to better recover the original image content, we conducted another experiment. The four state-of-the-art SR models are trained using LR images generated by the proposed CAR model trained jointly with the EDSR. Experimental results shown by the `CAR\textdaggerdbl' in Table \ref{tab:more_comparision} indicate that the performance of SR models trained using LR images generated by the CAR model trained jointly with the EDSR significantly surpasses that trained using images downscaled by the bicubic interpolation. We also observed that the performance gain of the CAR$\text{\textdaggerdbl}$ against the Bicubic is larger than the performance degradation against the CAR\textdagger. The two findings lead to the conclusion that the CAR image downscaling model does preserve content adaptive information that are essential to superior SR using deep SR models.
	
	\begin{figure*}[h]
		\centering
		\includegraphics[width=\textwidth]{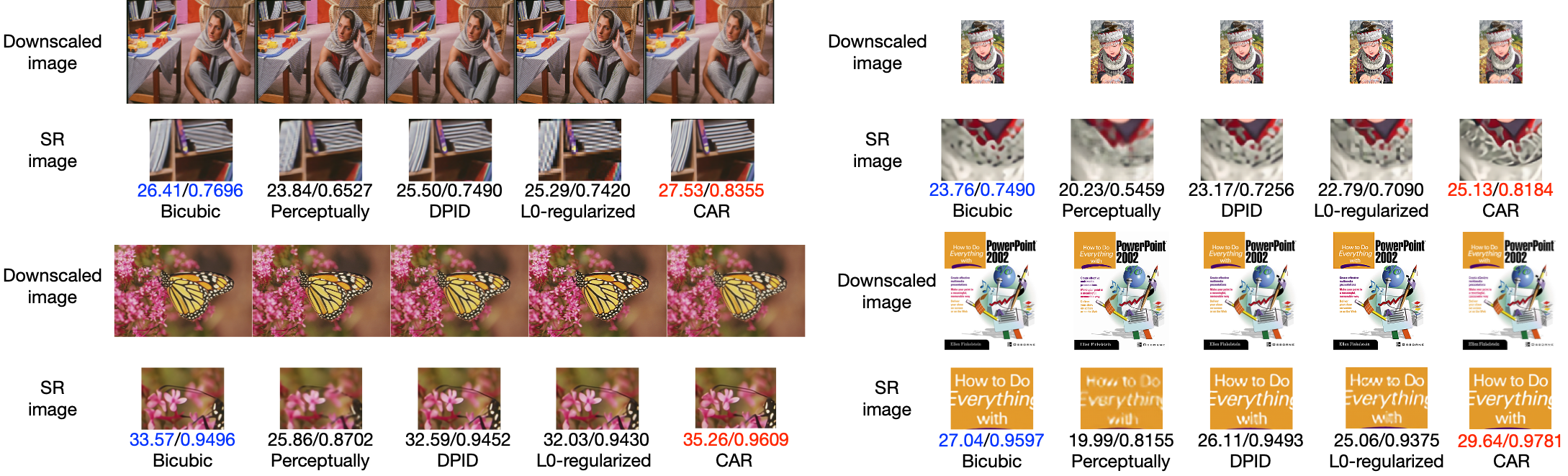}
		\medskip 
		\begin{minipage}{\textwidth} 
			{\footnotesize This figure is best viewed in color. Zoom in to see details of the downscaled image.\par}
		\end{minipage}
		\caption{Qualitative results of $4\times$ downscaled image and SR using the EDSR on four example images from the Set14 dataset. (More results are presented in the supplementary file.)}\label{fig:qualitative}
	\end{figure*}
	A qualitative comparison of $4\times$ downscaled images for SR is presented in Fig. \ref{fig:qualitative}. As can be seen, the CAR model produces downscaled images that are super-resolved with the best visual quality when compared with that of the other four models trained using the EDSR. As shown by the `Barbara' example, due to obvious aliasing occurred during downscaling by other four methods, the EDSR cannot recover the correct direction of the parallel edge pattern formed by a stack of books. The downscaled image generated by the CAR incurs less aliasing and the EDSR well recovered the direction of the parallel edge pattern. For the `Comic' example, we can observe that the SR result of the CAR downscaled image preserves more details. Visual results of the `PPT3' example demonstrates that the SR of downscaled images produced by the CAR better restore continuous edges and produce sharper HR images.
	
	\subsubsection{Ablation studies}
	\begin{table*}[h]
		\caption{Ablation results (PSNR / SSIM) of the CAR model on the Set5, Set14, BSD100, Urban100 and DIV2K (validation set).}\label{tab:ablation}
		\centering
		\resizebox{\textwidth}{!}{\begin{threeparttable}
			\begin{tabular}{c|c|c|c|c|c|c} 
				\hline
				Model                      & Scale & Set5                                                                        & Set14                                                                      & B100                                                                       & Urban100                                                                   & DIV2K                                                                       \\ 
				\hline\hline
				CAR                        & 2x    & \textcolor[rgb]{0.016,0.2,1}{38.94} / \textcolor[rgb]{0.016,0.2,1}{0.9658}  & \textcolor[rgb]{0.016,0.2,1}{35.61~}/ \textcolor[rgb]{0.016,0.2,1}{0.9404} & \textcolor[rgb]{0.016,0.2,1}{33.83}~/ \textcolor[rgb]{0.016,0.2,1}{0.9262} & \textcolor[rgb]{0.016,0.2,1}{35.24~}/ \textcolor[rgb]{0.016,0.2,1}{0.9572} & \textcolor[rgb]{0.016,0.2,1}{38.26} / \textcolor[rgb]{0.016,0.2,1}{0.9599}  \\
				CAR (w/o TV loss)          & 2x    & \textcolor[rgb]{1,0.125,0}{39.00~}/ \textcolor[rgb]{1,0.125,0}{0.9663}      & \textcolor[rgb]{1,0.125,0}{35.65~}/ \textcolor[rgb]{1,0.125,0}{0.9411}     & \textcolor[rgb]{1,0.125,0}{33.91~}/ \textcolor[rgb]{1,0.125,0}{0.9273}     & \textcolor[rgb]{1,0.125,0}{35.45} / \textcolor[rgb]{1,0.125,0}{0.9587}     & \textcolor[rgb]{1,0.125,0}{38.34} / \textcolor[rgb]{1,0.125,0}{0.9601}      \\
				CAR (w/o offset constrain) & 2x    & 38.73 / 0.9641                                                              & 35.29 / 0.9372                                                             & 33.58 / 0.9213                                                             & 35.19 / 0.9560                                                             & 38.03 / 0.9581                                                              \\
				CAR (w/o offset)           & 2x    & 38.09 / 0.9600                                                              & 34.46 / 0.9331                                                             & 33.25 / 0.9209                                                             & 32.91 / 0.9423                                                             & 37.24 / 0.9497                                                              \\ 
				\hline\hline
				CAR                        & 4x    & \textcolor[rgb]{0.016,0.2,1}{33.88~}/ \textcolor[rgb]{0.016,0.2,1}{0.9174}~ & \textcolor[rgb]{0.016,0.2,1}{30.31}~/ \textcolor[rgb]{0.016,0.2,1}{0.8382} & \textcolor[rgb]{0.016,0.2,1}{29.15~}/ \textcolor[rgb]{0.016,0.2,1}{0.8001} & \textcolor[rgb]{0.016,0.2,1}{29.28}~/ \textcolor[rgb]{0.016,0.2,1}{0.8711} & \textcolor[rgb]{0.016,0.2,1}{32.82~}/ \textcolor[rgb]{0.016,0.2,1}{0.8837}  \\
				CAR (w/o TV loss)          & 4x    & \textcolor[rgb]{1,0.125,0}{34.13} / \textcolor[rgb]{1,0.125,0}{0.9222}      & \textcolor[rgb]{1,0.125,0}{30.46} / \textcolor[rgb]{1,0.125,0}{0.8439}     & \textcolor[rgb]{1,0.125,0}{29.36~}/ \textcolor[rgb]{1,0.125,0}{0.8096}     & \textcolor[rgb]{1,0.125,0}{29.36} / \textcolor[rgb]{1,0.125,0}{0.8772}     & \textcolor[rgb]{1,0.125,0}{33.05} / \textcolor[rgb]{1,0.125,0}{0.8893}      \\
				CAR (w/o offset constrain) & 4x    & 33.86 / \textcolor[rgb]{0.016,0.2,1}{0.9174 }                               & 30.18 / 0.8368                                                             & 29.11 / 0.7998                                                             & 29.02 / 0.8704                                                             & 32.74 / 0.8835                                                              \\
				CAR (w/o offset)           & 4x    & 33.11 / 0.9168                                                              & 29.68 / 0.8322                                                             & 28.78 / 0.7921                                                             & 27.98 / 0.8675                                                             & 32.18 / 0.8830                                                              \\
				\hline
			\end{tabular}
			\begin{tablenotes}
				\small
				\item Note: \textcolor{red}{Red} color indicates the best performance and \textcolor{blue}{Blue} color represents the second.
			\end{tablenotes}
		\end{threeparttable}}
	\end{table*}
	We conducted ablation experiments on the proposed CAR model to verify the effectiveness of our design. We mainly concern about the contribution of kernel element offset and the constraint on offset distance to the performance of the SR. Table \ref{tab:ablation} shows the quantitative ablation results, from which we can observe that the SR performance on all testing cases constantly increases with the addition of kernel element offset and the constraint on offset distance. The baseline model is the CAR without kernel element offset (w/o offset), meaning that the CAR only needs to estimate the resampling kernel weights which will be applied to the position defined by Equation \ref{eqn:projection} on the HR image (also illustrated as the pixel center in Fig. \ref{fig:resampling}). Then, kernel element offset was incorporated, which brings a noticeable performance improvement. Introducing kernel element offset makes the resampling kernel to be non-uniform and each element in the resampling kernel can seek to proper sampling position to better preserve useful information for the end SR task. Further SR performance is gained by adding kernel element offset distance regularization. The kernel element offset distance regularization encourages the preservation of the resampling kernel topology and avoids unnecessary kernel element movement on the plain region with less structured texture, which potentially makes the training more stable and easier.
	
	\begin{figure}[h]
		\centering
		\subfigure[Input HR image]{\includegraphics[height=110px]{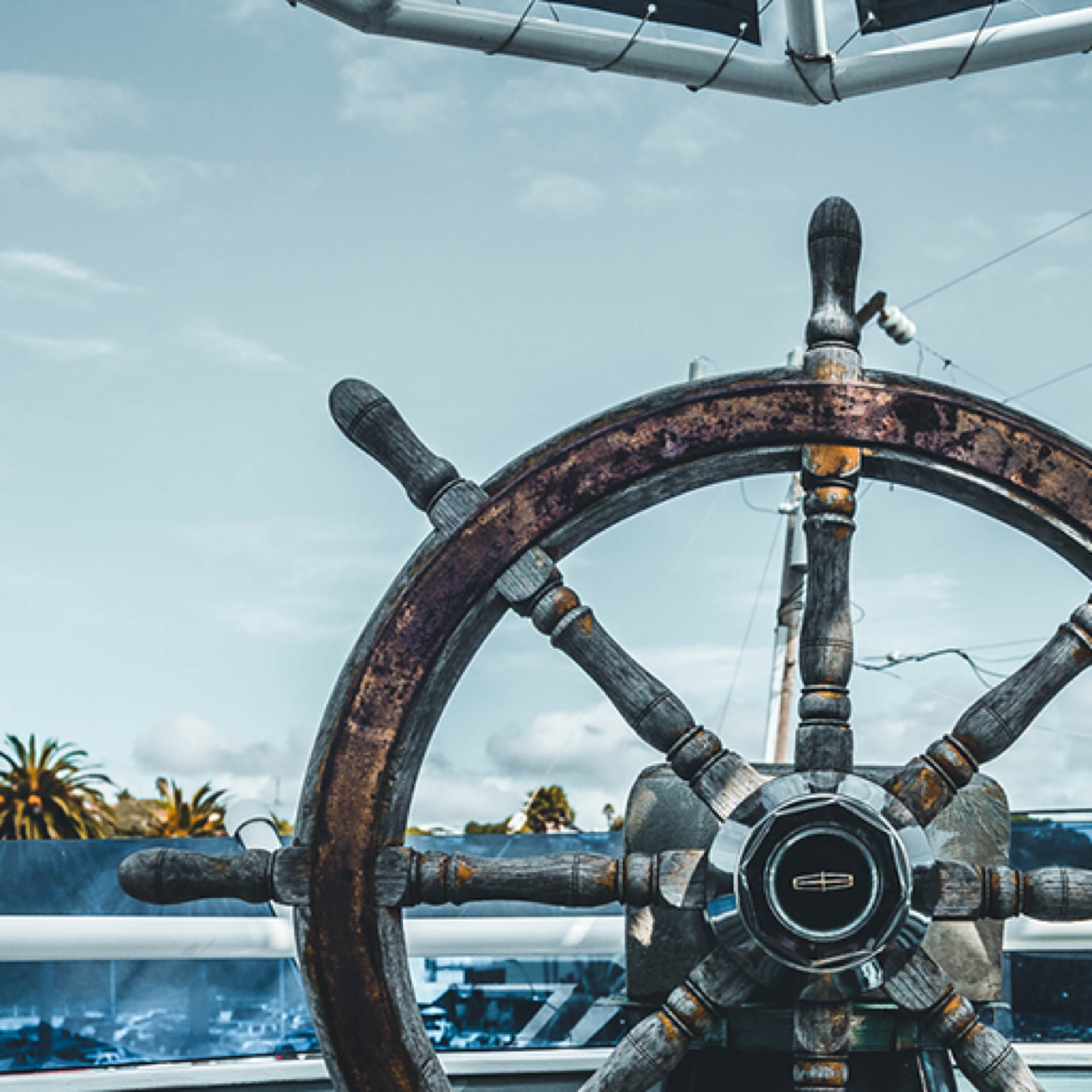}}\hfill
		\subfigure[Bicubic kernel]{\includegraphics[height=110px]{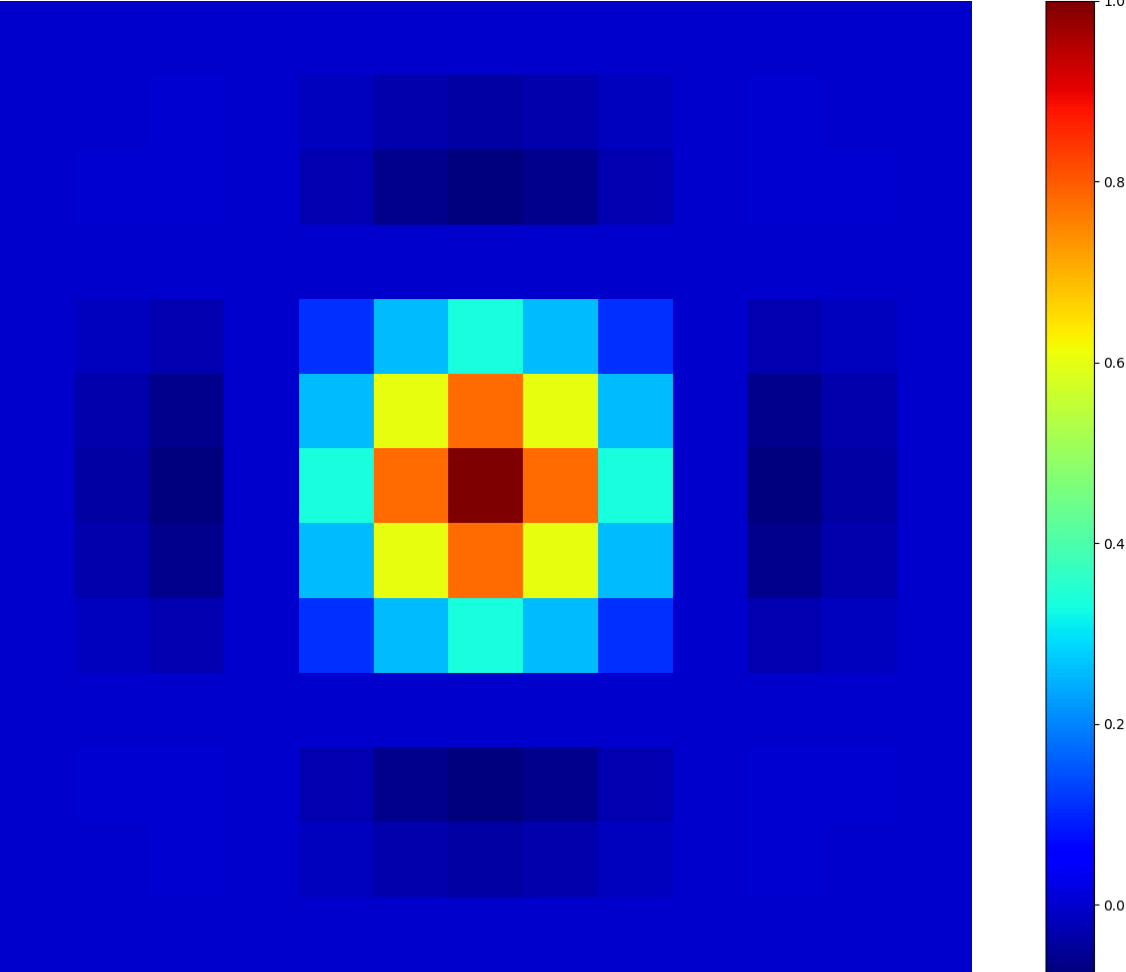}}\\
		\subfigure[CAR(kernels, w/o offset constraint)]{\includegraphics[height=110px]{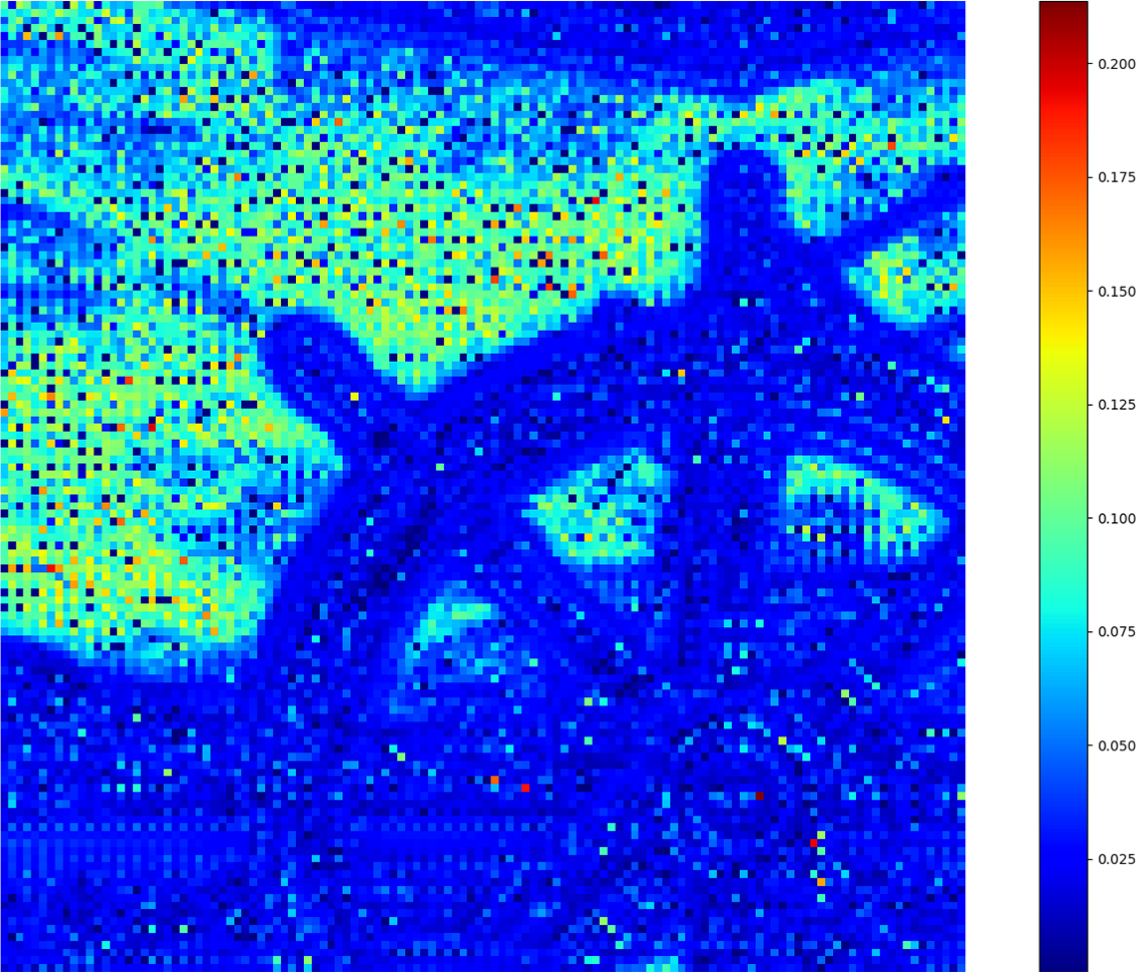}}\hfill
		\subfigure[CAR (offsets, w/o offset constraint)]{\includegraphics[height=110px]{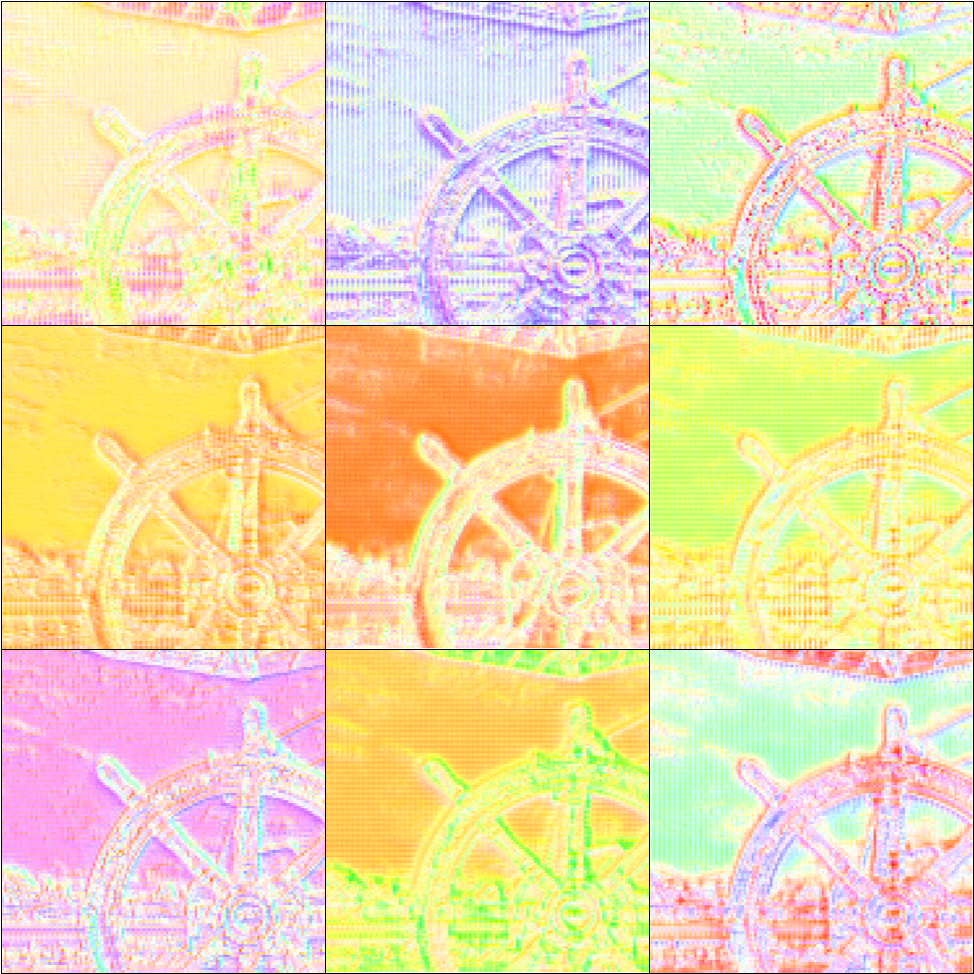}}\\
		\subfigure[kernels, w/ offset constraint]{\includegraphics[height=110px]{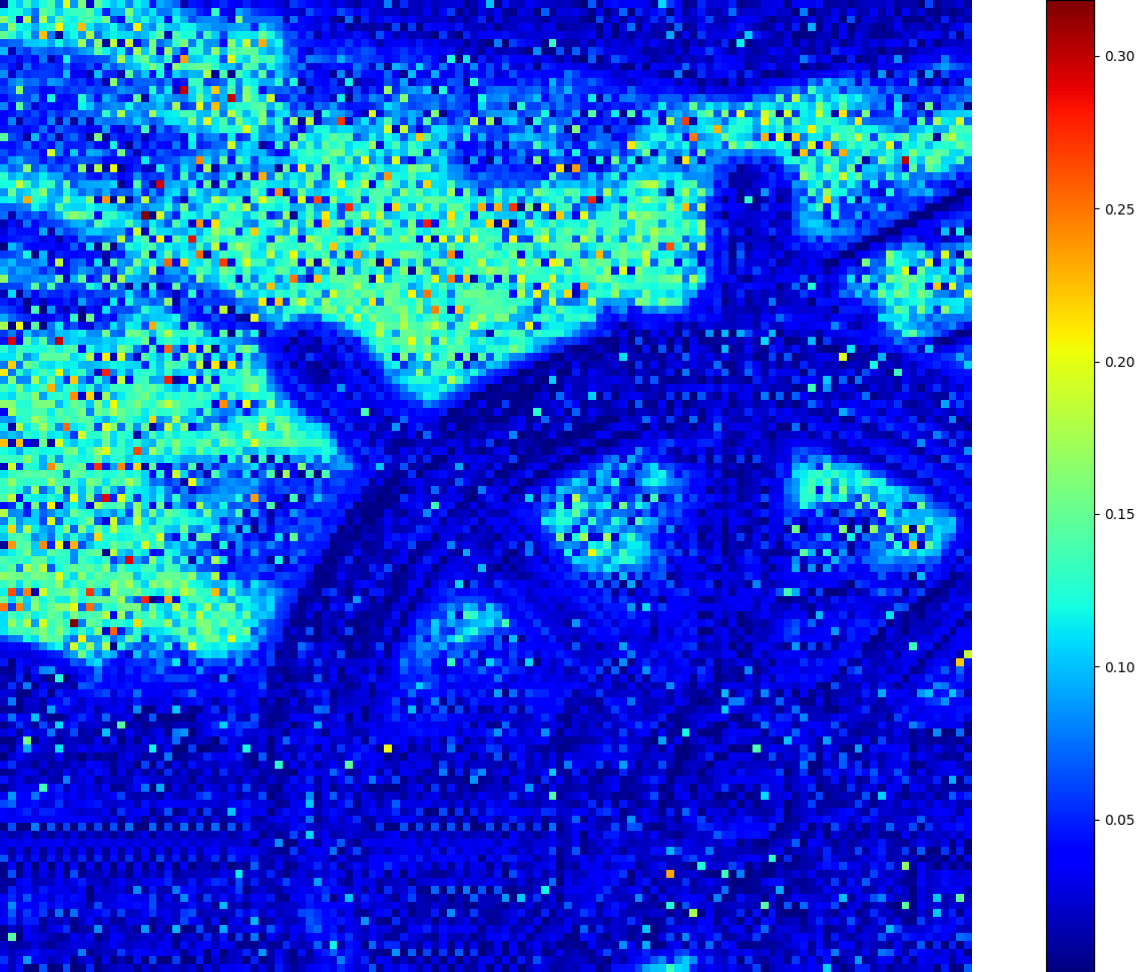}}\hfill
		\subfigure[CAR (offsets, w/ offset constraint)]{\includegraphics[height=110px]{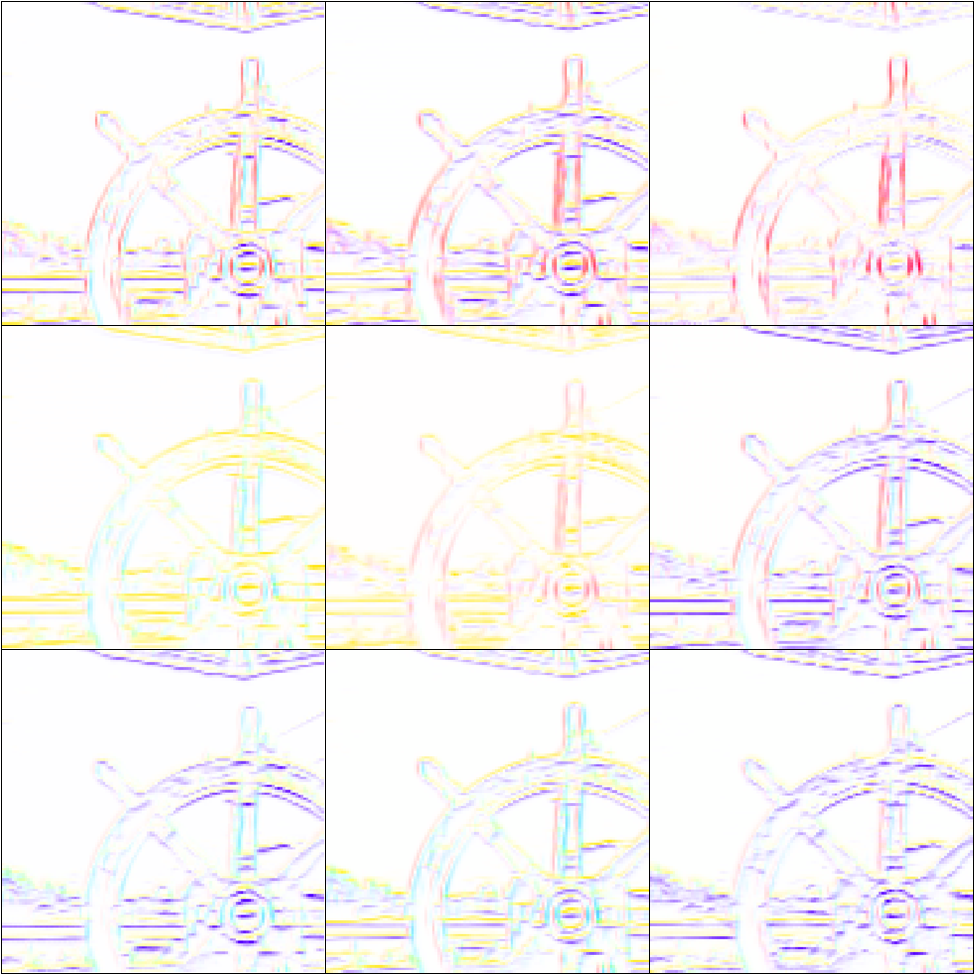}}
		\caption{An example of the $4\times$ downscaled resampling kernel elements and its corresponding offsets. In the figure we only visualized one bicubic resampling kernel, since it will be uniformly applied to all resampling regions. We only visualize the centeral 9 kernel elements ($(3\times h, 3\times w)$) predicted by the CAR model. \textcolor{blue}{The kernel element offset is visualized using the color wheel presented in \cite{baker:2007}.}}
		\label{fig:kernels}
	\end{figure}
	In order to better illustrate how the kernel offset distance regularization works, we visualized an example of resampling kernel elements and its corresponding offsets (Fig. \ref{fig:kernels}) in the configuration of with (w/) and without (w/o) the offset distance regularization. We only visualized the central 9 of many kernel elements and offsets for a better demonstration. Fig. \ref{fig:kernels} (c) and (d) present kernel elements and offsets by the CAR model trained without offset distance regularization. Fig. \ref{fig:kernels} (e) and (f) show the kernel elements and offsets estimated by the CAR model trained with offset distance regularization. It can be observed that kernel elements estimated by the CAR model trained with offset distance regularization only present obvious movement on the strong edges and textured regions (the wheel ring and handle), and almost hold still at the rather smoothed region (the sky region). The kernel elements estimated by the CAR model trained without offset distance regularization also move towards the strong edges. However, it presents intensive movements on the plain region, which may lead to an unstable training process and sub-optimal testing performance, since the gradient of the resampling kernel depends on the interpolated pixel value (Equation \ref{eqn:kernel_gradient}). The fixed bicubic kernel is also visualized in Fig. \ref{fig:kernels} (b). When compared with the content varying kernels shown in Fig. \ref{fig:kernels} (c) and (e), the fixed bicubic kernel will be uniformly applied to all the resampling region no matter what image content is going to be resampled.
	
	\begin{figure}[h]
		\centering
		\subfigure[HR patch]{\includegraphics[width=0.325\columnwidth]{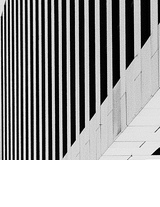}}\hfill
		\subfigure[w/o TV loss]{\includegraphics[width=0.325\columnwidth]{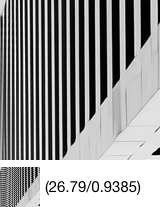}}\hfill
		\subfigure[w/ TV loss]{\includegraphics[width=0.325\columnwidth]{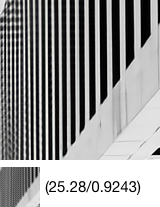}}
		\caption{An example trade off between perception of the $4\times$ downscaled image and distortion of the $4\times$ SR image. The first row shows the HR patch (a) and patch of SR results (b and c), the second row shows the downscaled version of the HR patch. The introduction of partial TV loss can produce visually better downscaled image at the expense of some SR performance.}
		\label{fig:aliasing_optimized}
	\end{figure}
	The superior SR performance with the CAR model is achieved from the powerful capability of deep neural networks that can approximate arbitrary functions. However, the deep learning model tends to find a tricky way to produce LR images preserving details that are in favor of generating accurate SR images but not for better human perception. Fig. \ref{fig:aliasing_optimized} shows an example of $4\times$ downscaled image by the CAR and $4\times$ SR image by the jointly trained EDSR. As shown in Fig. \ref{fig:aliasing_optimized} (b), the CAR model learned to preserve more information using much fewer pixel spaces by arranging vertical edges in a regular criss-cross way, which makes the vertical edges in the LR image look jaggy. Jaggies are one type of aliasing that normally manifest as regular artifacts near sharp changes in intensity. However, the human visual system finds regular artifacts more objectionable than irregular artifacts \cite{Cook:1986}. This problem is possibly caused by the inconsistent movement of the resampling kernels represented by the resampling kernel offsets near the sharp edges. To alleviate it, we introduced the partial TV loss of the horizontal and vertical resampling kernel offsets (Section \ref{sec:train_objective}) to constrain the rather free movements of the resampling kernel elements. As shown in Fig. \ref{fig:aliasing_optimized} (c), we can observe a smoother LR image with much less unsightly artifacts.
	
	By employing the partial TV loss of the resampling kernel offsets, the CAR model generates better images for perception. However, we also observed SR performance degradation on each testing dataset, which is shown in the `CAR' and `CAR (w/o TV loss)' entries of Table \ref{tab:ablation}. This is because the introduction of the partial TV loss breaks the optimal way of keeping information during downsampling. Other types of aliasing inevitably occurred on the sample-rate conversion, and the SR model cannot recover correct textures from those irregular patterns when compared with that of the regular jaggies. As illustrated by the SR patches shown in Fig. \ref{fig:aliasing_optimized} (b) and (c), we can recognize that the SR image corresponding the jagged LR image can better represent the original HR patch than that corresponding to the LR image with less jaggies.
	
	\begin{table*}[h]
		\caption{Public benchmark results (PSNR / SSIM) of 2$\times$ and 4$\times$ upscaling using different SR networks on the Set5, Set14, BSD100, Urban100 and DIV2K validation set.}\label{tab:compare_STOA}
		\centering
		\resizebox{\textwidth}{!}{\begin{threeparttable}
			\begin{tabular}{c|c|c|c|c|c|c|c|c} 
				\hline
				Donwscaling type          & Loss                & \multicolumn{2}{c|}{Upscaling}           & Set5                                                                       & Set14                                                                      & BSD100                                                                     & Urban100                                                                   & \multicolumn{1}{c}{DIV2K}                                                  \\ 
				\hline\hline
				\multirow{10}{*}{Bicubic} & \multirow{5}{*}{L2} & SRCNN             & \multirow{16}{*}{2x} & 36.66 / 0.9542                                                             & 32.42 / 0.9063                                                             & 31.36 / 0.8897                                                             & 29.50 / 0.8946                                                             & 33.05 / 0.9581                                                              \\
				&                     & VDSR              &                      & 37.53 / 0.9587                                                             & 33.03 / 0.9213                                                             & 31.90 / 0.8960                                                             & 30.76 / 0.9140                                                             & 33.66 / 0.9625                                                              \\
				&                     & DRRN              &                      & 37.74 / 0.9591                                                             & 33.23 / 0.9136                                                             & 32.05 / 0.8973                                                             & 31.23 / 0.9188                                                             & 35.63 / 0.9410                                                              \\
				&                     & MemNet            &                      & 37.78 / 0.9597                                                             & 33.28 / 0.9142                                                             & 32.08 / 0.8978                                                             & 31.31 / 0.9195                                                             & - / -                                                                       \\
				&                     & DnCNN             &                      & 37.58 / 0.9590                                                             & 33.03 / 0.9118                                                             & 31.90 / 0.8961                                                             & 30.74 / 0.9139                                                             & - / -                                                                       \\ 
				\cline{2-3}\cline{5-9}
				& \multirow{5}{*}{L1} & LapSRN            &                      & 37.52 / 0.9590                                                             & 33.08 / 0.9130                                                             & 31.80 / 0.8950                                                             & 30.41 / 0.9100                                                             & 35.31 / 0.9400                                                              \\
				&                     & ZSSR              &                      & 37.37 / 0.9570                                                             & 33.00 / 0.9108                                                             & 31.65 / 0.8920                                                             & - / -                                                                      & - / -                                                                       \\
				&                     & CARN              &                      & 37.76 / 0.9590                                                             & 33.52 / 0.9166                                                             & 32.09 / 0.8978                                                             & 31.92 / 0.9256                                                             & 36.04 / 0.9451                                                              \\
				&                     & SRRAM             &                      & 37.82 / 0.9592                                                             & 33.48 / 0.9171                                                             & 32.12 / 0.8983                                                             & 32.05 / 0.9264                                                             & - / -                                                                       \\
				&                     & ESRGAN            &                      & - / -                                                                      & - / -                                                                      & - / -                                                                      & - / -                                                                      & - / -                                                                       \\ 
				\cline{1-3}\cline{5-9}
				\multirow{6}{*}{Learned}  & \multirow{3}{*}{L2} & (CNN-CR)-(CNN-SR) &                      & 38.88 /~ ~-~ ~                                                             & 35.40 /~ ~-~ ~                                                             & \textcolor[rgb]{0.016,0.2,1}{33.92} /~ ~-~ ~                               & 33.68 /~ ~-~ ~                                                             & - / -                                                                       \\
				&                     & (CAR)-(CNN-SR)    &                      & 38.91 / 0.9656                                                             & 35.55 / 0.9401                                                             & \textcolor[rgb]{1,0.149,0}{33.96} / \textcolor[rgb]{1,0.149,0}{0.9281}     & 34.73 / 0.9539                                                             & 38.15 / 0.9593                                                              \\
				&                     & (CAR)-(EDSR)      &                      & \textcolor[rgb]{1,0.149,0}{39.01} / \textcolor[rgb]{1,0.149,0}{0.9662}     & \textcolor[rgb]{1,0.149,0}{35.64} / \textcolor[rgb]{1,0.149,0}{0.9406}     & 33.84 / \textcolor[rgb]{0.016,0.2,1}{0.9262}                               & \textcolor[rgb]{0.016,0.2,1}{35.15} / \textcolor[rgb]{0.016,0.2,1}{0.9568} & \textcolor[rgb]{0.016,0.2,1}{38.21} / \textcolor[rgb]{0.016,0.2,1}{0.9597}  \\ 
				\cline{2-3}\cline{5-9}
				& \multirow{3}{*}{L1} & (TAD)-(TAU)       &                      & 37.69 /~ ~-~ ~                                                             & 33.90 /~ ~-~ ~                                                             & 32.62 /~ ~-~ ~                                                             & 31.96 /~ ~-~ ~                                                             & 36.13 /~ ~-~ ~                                                              \\
				&                     & (CAR)-(TAU)       &                      & 37.93~/ 0.9628                                                             & 34.19~/ 0.9312                                                             & 32.78~/ 0.7592                                                             & 32.54~/ 0.9388                                                             & 36.86~/ 0.9524                                                              \\
				&                     & (CAR)-(EDSR)      &                      & \textcolor[rgb]{0.016,0.2,1}{38.94} / \textcolor[rgb]{0.016,0.2,1}{0.9658} & \textcolor[rgb]{0.016,0.2,1}{35.61} / \textcolor[rgb]{0.016,0.2,1}{0.9404} & 33.83 / \textcolor[rgb]{0.016,0.2,1}{0.9262}                               & \textcolor[rgb]{1,0.149,0}{35.24} / \textcolor[rgb]{1,0.149,0}{0.9572}     & \textcolor[rgb]{1,0.149,0}{38.26} / \textcolor[rgb]{1,0.149,0}{0.9599}      \\ 
				\hline\hline
				\multirow{11}{*}{Bicubic} & \multirow{6}{*}{L2} & SRCNN             & \multirow{17}{*}{4x} & 30.48 / 0.8628                                                             & 27.49 / 0.7503                                                             & 26.90 / 0.7101                                                             & 24.52 / 0.7221                                                             & 27.78 / 0.8753                                                              \\
				&                     & VDSR              &                      & 31.35 / 0.8838                                                             & 28.01 / 0.7674                                                             & 27.29 / 0.7251                                                             & 25.18 / 0.7524                                                             & 28.17 / 0.8841                                                              \\
				&                     & SRResNet          &                      & 32.05 / 0.8910                                                             & 28.53 / 0.7804                                                             & 27.57 / 0.7354                                                             & 26.07 / 0.7839                                                             & - / -                                                                       \\
				&                     & DRRN              &                      & 31.68 / 0.8888                                                             & 28.21 / 0.7720                                                             & 27.38 / 0.7284                                                             & 25.44 / 0.7638                                                             & 29.98 / 0.8270                                                              \\
				&                     & MemNet            &                      & 31.74 / 0.8893                                                             & 28.26 / 0.7723                                                             & 27.40 / 0.7281                                                             & 25.50 / 0.7630                                                             & - / -                                                                       \\
				&                     & DnCNN             &                      & 31.40 / 0.8845                                                             & 28.04 / 0.7672                                                             & 27.29 / 0.7253                                                             & 25.20 / 0.7521                                                             & - / -                                                                       \\ 
				\cline{2-3}\cline{5-9}
				& \multirow{5}{*}{L1} & LapSRN            &                      & 31.54 / 0.8850                                                             & 28.19 / 0.7720                                                             & 27.32 / 0.7280                                                             & 25.21 / 0.7560                                                             & 29.88 / 0.8250                                                              \\
				&                     & ZSSR              &                      & 31.13 / 0.8796                                                             & 28.01 / 0.7651                                                             & 27.12 / 0.7211                                                             & - / -                                                                      & - / -                                                                       \\
				&                     & CARN              &                      & 32.13 / 0.8937                                                             & 28.60 / 0.7806                                                             & 27.58 / 0.7349                                                             & 26.07 / 0.7837                                                             & 30.43 / 0.8374                                                              \\
				&                     & SRRAM             &                      & 32.13 / 0.8932                                                             & 28.54 / 0.7800                                                             & 27.56 / 0.7350                                                             & 26.05 / 0.7834                                                             & - / -                                                                       \\
				&                     & ESRGAN            &                      & 32.73 / 0.9011                                                             & 28.99 / 0.7917                                                             & 27.85 / 0.7455                                                             & 27.03 / 0.8153                                                             & - / -                                                                       \\ 
				\cline{1-3}\cline{5-9}
				\multirow{6}{*}{Learned}  & \multirow{3}{*}{L2} & (CNN-CR)-(CNN-SR) &                      & - / -                                                                      & - / -                                                                      & - / -                                                                      & - / -                                                                      & - / -                                                                       \\
				&                     & (CAR)-(CNN-SR)    &                      & 33.73 / 0.9148                                                             & 30.22 / 0.8329                                                             & 29.13 / 0.7992                                                             & 28.56 / 0.8546                                                             & 32.67 / 0.8800                                                              \\
				&                     & (CAR)-(EDSR)      &                      & \textcolor[rgb]{0.016,0.2,1}{33.87} / \textcolor[rgb]{1,0.149,0}{0.9176}   & \textcolor[rgb]{1,0.149,0}{30.34} / \textcolor[rgb]{0.016,0.2,1}{0.8381}   & \textcolor[rgb]{1,0.149,0}{29.18} / \textcolor[rgb]{1,0.149,0}{0.8005}     & \textcolor[rgb]{0.016,0.2,1}{29.23} / \textcolor[rgb]{0.016,0.2,1}{0.8710} & \textcolor[rgb]{1,0.149,0}{32.85} / \textcolor[rgb]{0.016,0.2,1}{0.8835}    \\ 
				\cline{2-3}\cline{5-9}
				& \multirow{3}{*}{L1} & (TAD)-(TAU)       &                      & 31.59 /~ ~-~ ~                                                             & 28.36 /~ ~-~ ~                                                             & 27.57 /~ ~-~ ~                                                             & 25.56 /~ ~-~ ~                                                             & 30.25 /~ ~-~ ~                                                              \\
				&                     & (CAR)-(TAU)       &                      & 31.85 / 0.8938                                                             & 28.77 / 0.8028                                                             & 27.84 / 0.7592                                                             & 26.15 / 0.7978                                                             & 31.10 / 0.8512                                                              \\
				&                     & (CAR)-(EDSR)      &                      & \textcolor[rgb]{1,0.149,0}{33.88} / \textcolor[rgb]{0.016,0.2,1}{0.9174}   & \textcolor[rgb]{0.016,0.2,1}{30.31} / \textcolor[rgb]{1,0.149,0}{0.8382}   & \textcolor[rgb]{0.016,0.2,1}{29.15} / \textcolor[rgb]{0.016,0.2,1}{0.8001} & \textcolor[rgb]{1,0.149,0}{29.28} / \textcolor[rgb]{1,0.149,0}{0.8711}     & \textcolor[rgb]{0.016,0.2,1}{32.82} / \textcolor[rgb]{1,0.149,0}{0.8837}    \\
				\hline
			\end{tabular}
			\begin{tablenotes}
				\small
				\item Note: \textcolor{red}{Red} color indicates the best performance and \textcolor{blue}{Blue} color represents the second. The `-' indicates that results are not provided by the corresponding original publication.
			\end{tablenotes}
		\end{threeparttable}}
	\end{table*}
	\subsubsection{Comparison with public benchmark SR results}
	To compare the SR performance of downscaled images generated by the proposed CAR model with other deep SR models trained neither on images downscaled using predefined operations (such as bicubic interpolation) or images downscaled by end-to-end trained task driven image downscaling models, we listed several commonly compared public benchmark results on Table \ref{tab:compare_STOA}. The comparison was organized into two groups, SR models trained with LR images produced using the bicubic interpolation method (`Bicubic` group) and LR images generated by models trained under the guidance of SR task (`Learned' group). In each group, it was further split into models trained using L2 loss function and L1 loss function. SR models in the `Bicubic' group include the SRCNN \cite{dong:2016}, VDSR \cite{kim:2016}, DRRN \cite{tai:2017}, MemNet \cite{chen:2018}, DnCNN \cite{zhangkai:2017}, LapSRN \cite{lai:2018}, ZSSR \cite{shcocher:2018}, CARN \cite{ahn:2018}, SRRAM \cite{kim:2018ram}. In the `Learned' group, we compared the CAR model with two recent state-of-the-art image downscaling models trained jointly with deep SR models, \ie, the CNN-CR$\rightarrow$CNN-SR \cite{li:2019} model and the TAD$\rightarrow$TAU model \cite{kim:2018}. We also jointly trained the CAR model with the CNN-SR and TAU to do fair comparisons with the performance reported in its corresponding paper.
	
	As shown in Table \ref{tab:compare_STOA}, SR performance of upscaling models trained jointly with learnable image downscaling model outperform that of SR models trained using images downscaled in a predefined manner. This is because that LR images generated by learnable image downscaling models adaptively preserve content dependent information during downscaling, which essentially helps deep SR models learn to better recover original image content. Comparing our model with recent state-of-the-art SR driven image downscaling models, the proposed model achieved the best PSNR performance on four out of five testing datasets on the 2$\times$ image downscaling and upscaling track. On the 4$\times$ track, the proposed model achieved the best performance. Noticeable improvement on the Urban100 dataset \cite{huang:2015} can be observed. This can be possibly explained by the reason that images of the Urban100 are all buildings with rich sharp edges, and TAD is trained under the constraint of producing LR images similar to images downscaled by bicubic interpolation with pre-filtering which inevitably constrain the edges of LR images generated by the TAD to be blurry like that downscaled by the bicubic interpolation. Therefore, the jointly trained up sampler cannot well recover original edges from those blurred edges of the LR images. As to the proposed CAR model, it is trained without any LR constraint, since the resampling method naturally makes the LR result a valid image. Benefiting from the unsupervised training strategy, the CAR model can adaptively preserve essential edge information for better super-resolution.
	
	\subsection{Evaluation of downscaled images}
\begin{figure}[h]
		\centering
		\subfigure[HR image]{\includegraphics[width=0.32\columnwidth]{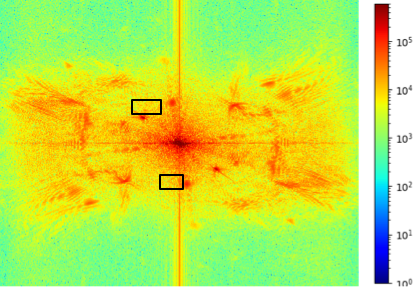}}
		\subfigure[Bicubic]{\includegraphics[width=0.32\columnwidth]{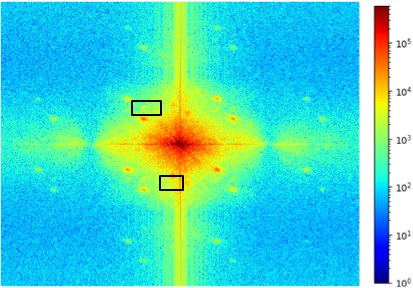}}
		\subfigure[Perceptually]{\includegraphics[width=0.32\columnwidth]{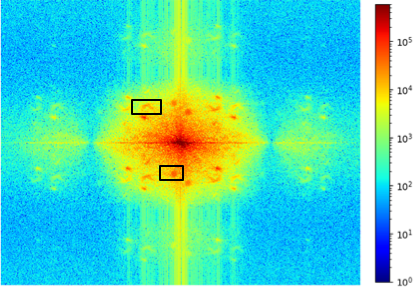}}\\
		\subfigure[DPID]{\includegraphics[width=0.32\columnwidth]{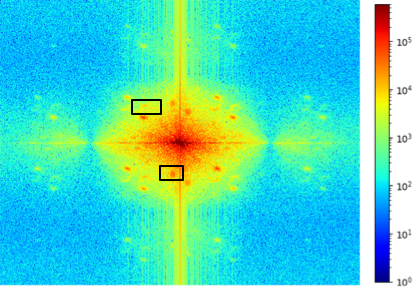}}
		\subfigure[L0-regularized]{\includegraphics[width=0.32\columnwidth]{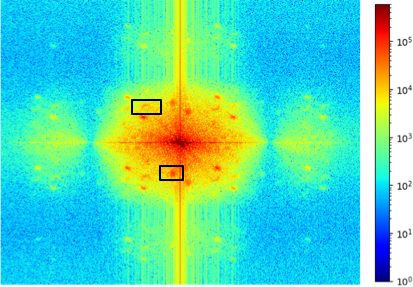}}
		\subfigure[CAR]{\includegraphics[width=0.32\columnwidth]{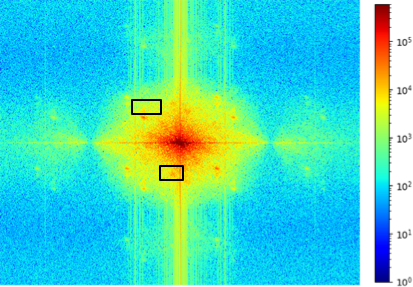}}\\
		\subfigure[Cropped patch from the downscaled 'Barbara' image]{\includegraphics[width=0.96\columnwidth]{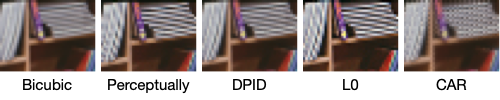}}
		\caption{Spectrum analysis of the 4$\times$ downscaled `Barbara' image in the Set14 dataset using different downscaling methods.}
		\label{fig:fft}
	\end{figure}

	This section presents the analysis of the quality of the downscaled image produced by the CAR model from two perspectives. We first analyzed the downscaled image in the frequency domain. We used the `Barbara' image from the Set14 dataset as an example because it contains a wealth of high-frequency components (as shown in the first example of Fig. \ref{fig:qualitative}). Fig. \ref{fig:fft} shows the spectrum obtained by applying FFT on the HR image and the downscaled images generated by the CAR model and four downscaling methods been compared. Each point in the spectrum represents a particular frequency contained in the spatial domain image. The point in the center of the spectrum is the DC component, and points closely around the center point represents low-frequency components. The further away from the center a point is, the higher is its corresponding frequency.
	
	The spectrum of the HR image (Fig. \ref{fig:fft} (a)) contains a lot of high-frequency components. During image downscaling, spatial-aliasing is inevitably occurred since the sampling rate is below the Nyquist frequency. Aliasing can be spotted in the spectrum as spurious bands that are not presented in the spectrum of the HR image (high frequency component is aliased into low frequency), \eg, the black box marked regions in Fig. \ref{fig:fft} (c-e) compared to thaty of the original spectrum in Fig. \ref{fig:fft} (a). One way to remove aliasing is to use a blurry filter upon resampling (so does the default MATLAB \texttt{imresize} function). As shown by the black box marked regions in Fig. \ref{fig:fft} (b), aliasing of the downscaled image produced by the MATLAB \texttt{imresize} function is alleviated. Compared with the spectrum of image produced by the three state-of-the-art image downscaling methods (Fig. \ref{fig:fft} (c-e)), less power exaggeration of low frequency components can be observed in the black box indicated regions. Therefore, the spectrum of the CAR generated image demonstrates that less aliasing is produced during the downscaling by the CAR model. A cropped region from the downscaled `Barbara' image is shown in Fig. \ref{fig:fft} (g) and less spatial aliasing can be observed from the Bicubic and CAR downscaled image when compared with that of the other three image downscaling methods.
	
	\begin{table}[h]
		\centering
		\caption{Bpp of lossless compressed downscaled images using the JPEG-LS}\label{tab:bpp}
		\resizebox{\columnwidth}{!}{\begin{threeparttable}
			\begin{tabular}{c|c|ccccc} 
				\hline
				\multicolumn{1}{c}{} & \multicolumn{1}{c}{} & Bicubic                               & Perceptually & DPID   & L$_0$  & CAR                                  \\ 
				\hline\hline
				Set5                 & \multirow{6}{*}{2x}  & \textcolor[rgb]{0.016,0.2,1}{11.99}   & 13.27        & 12.65  & 16.10  & \textcolor[rgb]{1,0.145,0}{11.80}    \\
				Set14                &                      & \textcolor[rgb]{0.016,0.2,1}{11.23}   & 12.01        & 12.05  & 15.07  & \textcolor[rgb]{1,0.145,0}{11.14}    \\
				BSD100               &                      & \textcolor[rgb]{0.016,0.2,1}{11.11}   & 11.90        & 11.8   & 15.31  & \textcolor[rgb]{1,0.145,0}{10.69}    \\
				Urban100             &                      & \textcolor[rgb]{0.016,0.2,1}{11.39}   & 12.22        & 12.19  & 15.27  & \textcolor[rgb]{1,0.145,0}{11.28}    \\
				DIV2K                &                      & \textcolor[rgb]{0.016,0.2,1}{10.42}   & 11.38        & 11.1   & 14.04  & \textcolor[rgb]{1,0.145,0}{10.20}    \\
				Average              &                      & \textcolor[rgb]{0.016,0.2,1}{11.228}  & 12.156       & 11.958 & 15.158 & \textcolor[rgb]{1,0.145,0}{11.022}   \\ 
				\hline\hline
				Set5                 & \multirow{6}{*}{4x}  & \textcolor[rgb]{0.016,0.2,1}{14.78}   & 16.52        & 15.32  & 17.41  & \textcolor[rgb]{1,0.145,0}{14.39}    \\
				Set14                &                      & \textcolor[rgb]{0.016,0.2,1}{12.61}   & 14.27        & 13.32  & 15.19  & \textcolor[rgb]{1,0.145,0}{12.47}    \\
				BSD100               &                      & \textcolor[rgb]{0.016,0.2,1}{12.85}   & 14.42        & 13.51  & 15.59  & \textcolor[rgb]{1,0.145,0}{12.61}    \\
				Urban100             &                      & \textcolor[rgb]{1,0.145,0}{12.23}     & 14.26        & 12.94  & 14.97  & \textcolor[rgb]{0.016,0.2,1}{12.28}  \\
				DIV2K                &                      & \textcolor[rgb]{0.016,0.2,1}{11.54}   & 13.46        & 12.14  & 13.91  & \textcolor[rgb]{1,0.145,0}{11.39}    \\
				Average              &                      & \textcolor[rgb]{0.016,0.2,1}{12.802}  & 14.586       & 13.446 & 15.414 & \textcolor[rgb]{1,0.145,0}{12.628}   \\
				\hline
			\end{tabular}
			\begin{tablenotes}
				\small
				\item Note: \textcolor{red}{Red} color indicates the best performance and \textcolor{blue}{Blue} color represents the second.
			\end{tablenotes}
		\end{threeparttable}}
	\end{table}
	Additionally, we analyzed the downscaled image from the perspective of lossless compression. Table \ref{tab:bpp} presents the average bits-per-pixel (bpp) of the lossless compressed images using the lossless JPEG (JPEG-LS) \cite{weinberger:2000}. Downscaled images produced by the CAR model can be more easily compressed than that by downscaling algorithms designed for better human perception. The reason why this happened is that downscaling methods designed for better human perception tend to preserve or enhance edges in a pre-defined manner, and much more high frequency components are retained, thus the downscaled images are less compressible on average. When compared with the bpp of compressed bicubic downscaled images, the CAR model achieved similar compression performance.
	
	\subsection{User study}
	\begin{figure}[h]
		\centering
		\includegraphics[width=\columnwidth]{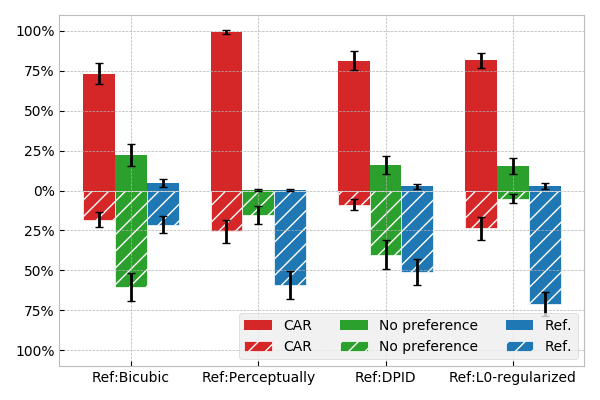}
		\caption{User study results comparing the proposed CAR model against reference image downscaling algorithms for the image super-resolution and image downscaling tasks. Each data point is an average over valid records evaluated on 20 image groups and the error bars indicate 95\% confidence interval. The upper part (above the zero axes) is the image quality preference of the SR task, and the lower part (below the zero axes) is the image quality preference of the image downscaling task.}
		\label{fig:mos}
	\end{figure}
	To evaluate the visual quality of the generated SR images and LR images corresponding to different downscaling algorithms, we conducted user study which is widely adopted in many image generation tasks. We picked 59 sample images from different testing datasets, \ie, the Set5 (5 images), Set14 (20 images), BSD100 (20 images), and Urban100 (20 images) dataset. Samples are randomly selected with diverse properties, including people, animal, building, natural scenes and computer-generated graphics. We adopted similar evaluation settings used in \cite{kopf:2013,oztireli:2015,weber:2016,liu:2018}. The user study were conducted as the A/B testing: the original image was presented in the middle place with two variants (SR images or downscaled images) showed in either side, among which one is produced by the CAR method and another is generated by one of the competing methods, \ie, Bicubic, Perceptually \cite{oztireli:2015}, DPID \cite{weber:2016} and L0-regularized \cite{liu:2018}. Users were required to answer the question `which one looks better' by exclusively selecting one of the three options from: 1) A is better than B; 2) A equals to B; 3) B is better than A. For all 59 sample images, there are 236 pairwise decisions for each user. All image pairs were shown in random temporal order and the two variants of the original image of a pair are also randomly shown in position A or position B. To test the reliability of the user study result, we add additional 10 image groups by randomly repeating question from the 236 trials.  All images were displayed at native resolution of the monitor and zoom functionality of the UI was disabled. Users can only pan the view if the total size of a group of images exceeds the screen resolution. No other restrictions on the viewing way were imposed and users can judge those images at any viewing distance and angle without time limits.
	
	We invited 29 participants for both the user study on $4\times$ image super-resolution and $4\times$ image downscaling tasks. Answers from each participant are filtered out if it achieves less than 80\% consistency \cite{kopf:2013,oztireli:2015} on the repeated 10 questions, which generates 29 and 28 valid records for the image super-resolution and image downscaling tasks, respectively. As can be seen from Fig. \ref{fig:mos}, the total preferences of the SR and image downscaling task opted for the CAR model are larger than that of the reference methods.  The upper part (above the zero axes) of Fig. \ref{fig:mos} shows the results of the user study on $4\times$ image super-resolution task, each group of bars present the comparison between SR images corresponding to the CAR generated LR images and SR images corresponding to LR images produced by the reference method. The results indicate that the CAR model achieves at least 73\% preference over all other algorithms. Besides, our algorithm achieves more than 98\% preference compared with the Perceptually method, demonstrating that there is a distinct difference between the SR image corresponding to the perceptually based \cite{oztireli:2015} downscaled image and the original HR image. Although there are about 20\% preference on `A equals to B' plus `B is better than A' on the DPID and L0-regularized entry, it still cannot compete with the significant superiority of the CAR method. When combined with the objective metrics presented in Table \ref{tab:comparision}, we can arrive at the conclusion that LR images produced by algorithms solely optimized for better human perception cannot be well recovered.
	
	The lower part (below the zero axes) of Fig. \ref{fig:mos} shows the users' perceptual preference for the $4\times$ downscaled images generated by the CAR method versus the Bicubic, Perceptually, DPID, and L0-regularized image downscaling algorithms. We observed that the CAR method gets distinct less preference when compared with that of the Perceptually, DPID, and L0-regularized algorithm, which illustrates that the three state-of-the-art algorithms generate more perceptually favored LR images. However, when compared with the Bicubic downscaling algorithm, the user preference for the CAR method is slightly inferior to that for the Bicubic, and at most cases, participants tend to give no preference for both methods. This indicates that the CAR image downscaling method is comparable to the bicubic downscaling algorithm in terms of human perception. Among the three state-of-the-art image downscaling algorithms, the DPID achieves less agreement since its hyper-parameter is content dependent, and during the test, default value is used. The perceptually based image downscaling and the L0-regularized method achieve more than 75\% user preference because these methods artificially emphasize the edges of image content, which can be good if the image is going to be displayed at a very small size, such as an icon. Whether it is desirable in other cases is debatable. It does tend to make images look better at first glance, but at the expense of realism in terms of signal fidelity.
	
	\section{Conclusion}\label{sec:conclusion}
	This paper introduces the CAR model for image downscaling, which is an end-to-end system trained by maximizing the SR performance. It simultaneously learns a mapping for resolution reduction and SR performance improvement. One major contribution of our work is that the CAR model is trained in an unsupervised manner meaning that there is no assumption on how the original HR image will be downscaled, which helps the image downscale model to learn to keep essential information for SR task in a more optimal way. This is achieved by the content adaptive resampling kernel generation network which estimates spatial non-uniform resampling kernels for each pixel in the downscaled image according to the input HR image. The downscaled pixel value is obtained by decimating HR pixels covered by the resampling kernel. Our experimental results illustrate that the CAR model trained jointly with the SR networks achieves a new state-of-the-art SR performance while produces downscaled images whose quality are comparable to that of the widely adopted image downscaling method.



\normalsize
\bibliography{main}


\end{document}